\RequirePackage{fix-cm}
\documentclass[twocolumn]{svjour3}          % twocolumn
\smartqed  % flush right qed marks, e.g. at end of proof
\usepackage{graphicx}
\usepackage[numbers]{natbib}
\usepackage{times}
\usepackage{epsfig}
\usepackage{amsmath}
\usepackage{amssymb}
\usepackage{soul}
\usepackage{dsfont}
\usepackage{multirow}
\usepackage{capt-of}
\usepackage{graphics}
\usepackage{algorithm}
\usepackage{algorithmic}
\usepackage{dsfont}
\usepackage{booktabs} 

\newcommand{\mb}[1]{\mathbf{#1}}

\begin{document}

\title{Complete 3D Scene Parsing from an RGBD Image%\thanks{Grants or other notes
%about the article that should go on the front page should be
%placed here. General acknowledgments should be placed at the end of the article.}
}
%\subtitle{Do you have a subtitle?\\ If so, write it here}

%\titlerunning{Short form of title}        % if too long for running head

\author{Chuhang Zou         \and
        Ruiqi Guo \and Zhizhong Li \and Derek Hoiem%etc.
}

%\authorrunning{Short form of author list} % if too long for running head

\institute{C. Zou$^1$ \and  R. Guo$^2$ \and Z. Li$^1$ \and  D. Hoiem$^1$\at
              $^1$Department of Computer Science, University of Illinois\\ at Urbana-Champaign, Champaign, USA \\
              \email{\{czou4, zli115, dhoiem\}@illinois.edu} \\
              $^2$Google Reserch\\
              \email{guorq@google.com}
              %  \\
%             \emph{Present address:} of F. Author  %  if needed
}

\date{Received: date / Accepted: date}
% The correct dates will be entered by the editor

\maketitle

\begin{abstract}
One major goal of vision is to infer physical models of objects, surfaces, and their layout from sensors. In this paper, we aim to interpret indoor scenes from one RGBD image. Our representation encodes the layout of orthogonal walls %which must conform to a Manhattan structure 
and the extent of objects, modeled with CAD-like 3D shapes. We parse both the visible and occluded portions of the scene and all observable objects, producing a $complete$ 3D parse. Such a scene interpretation is useful for robotics and visual reasoning, but difficult to produce due to the well-known challenge of segmentation, the high degree of occlusion, and the diversity of objects in indoor scenes. We take a data-driven approach, generating sets of potential object regions, matching to regions in training images, and transferring and aligning associated 3D models while encouraging fit to observations and spatial consistency. We use support inference to aid interpretation and propose a retrieval scheme that uses convolutional neural networks (CNNs) to classify regions and retrieve objects with similar shapes. We demonstrate the performance of our method on our newly annotated NYUd v2 dataset~\cite{silberman2012indoor} with detailed 3D shapes.%, comparing with prior work and sensible baselines.
% \PACS{PACS code1 \and PACS code2 \and more}
% \subclass{MSC code1 \and MSC code2 \and more}
\end{abstract}

\begin{figure}[!ht]
\centering
\includegraphics[width=\columnwidth]{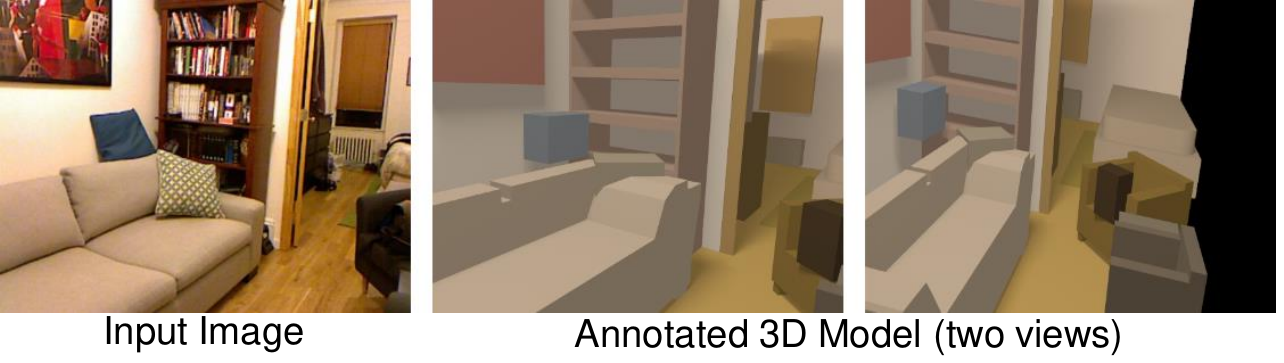}
%\vspace{-0.25in}
\caption{
\label{fig:intro}
Our goal is to recover a 3D model (right) from a single RGBD image (left), consisting of the position, orientation, and extent of layout surfaces and objects.}
\vspace{-1em}
\end{figure}

\begin{figure*}[t]
\begin{center}
%\fbox{\rule{0pt}{1.5in} \rule{0.9\linewidth}{0pt}}
   \includegraphics[width=0.95\linewidth]{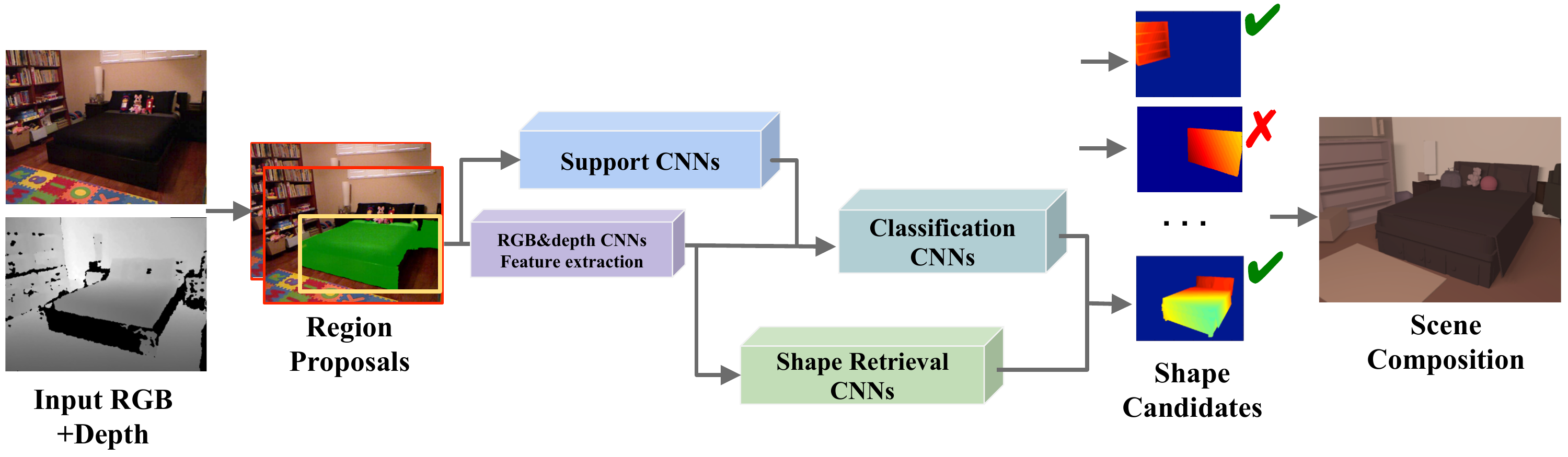}
\end{center}
\vspace{-1em}
   \caption{\textbf{Overview of our approach.} Given an input RGB-D (left), we propose possible layouts and object regions. We predict each object proposal's support height and class by our support and classification CNNs. We then retrieve a similar object shape, by using our shape retrieval CNNs to match the object region to the most similar region in the training set and transferring and aligning associated 3D models to the input depth image. The subset of proposed objects and layouts are then optimally selected based on consistency with observed depth, coverage, and constraints on occupied space. We show an example result (upper-right) and ground truth annotations (lower-right).}
   \vspace{-1em}
\label{fig:illustration}
\end{figure*}

\section{Introduction}
\label{intro}
Recovering the layout and shape of surfaces and objects is a foundational problem in computer vision.
Early approaches, such as reconstruction from line drawings~\cite{roberts1963machine}, attempt to infer 3D object and surface models based on shading cues or boundary reasoning. But the complexity of natural scenes is too difficult to model with hand-coded processing and rules.  More recent approaches to 3D reconstruction produce detailed literal geometric models, such as 3D point clouds or meshes, from multiple images~\cite{furukawa2009iccv}, or coarse interpreted models, such as boxy objects within a box-shaped room~\cite{hedau2009iccv}, from one image.

This paper introduces an approach to recover complete 3D models of indoor objects and layout surfaces from an RGBD (RGB+Depth) image (Fig.~\ref{fig:intro}). %, aiming to bridge the gap between detailed literal and coarse interpretive 3D models. 
Recovering 3D models from images is highly challenging due to three ambiguities: the loss of depth information when points are projected onto an image; the loss of full 3D geometry due to occlusion; and the unknown separability of objects and surfaces.  In this paper, we choose to work with RGBD images, rather than RGB images, so that we can focus on designing and inferring a useful representation, without immediately struggling with the added difficulty of interpreting geometry of visible surfaces.  Even so, ambiguities due to occlusion and unknown separability of nearby objects make 3D reconstruction impossible in the abstract.  If we see a book on a table, we observe only part of the book's and table's surfaces.  How do we know their full shape, or even that the book is not just a bump on the table?  Or how do we know that a sofa does not occlude a hole in the wall or floor?  We don't.  But we expect rooms to be enclosed, typically by orthogonal walls and horizontal surfaces, and we can guess the extent of the objects based on experience with similar objects.  Our goal is to provide computers with this same interpretive ability.

\textbf{Scene representation.} Ideally, we want an expressive representation that supports robotics (e.g., where is it, what is it, how to move and interact with it), graphics (e.g., what would the scene look like with or without this object), and interpretation (e.g., what is the person trying to do).  Importantly, we want to understand what is in the scene and what could be done, rather than only a labeling of visible surfaces. In this paper, we aim to infer a 3D geometric model that encodes the position and extent of layout surfaces, such as walls and floor, and objects such as tables, chairs, mugs, and televisions. In the long term, we hope to augment this geometric model with relations (e.g., this table supports that mug) and attributes (e.g., this is a cup that can serve as a container for small objects and can be grasped this way). %Our goal is to automatically determine the position, orientation, and extent of layouts and objects. 
%Our long-term goal is to incorporate all of these facets of scene understanding: parsing into objects, estimating extent, support reasoning, categorization, affordance, detailed shape estimation, etc.  In this paper, we take a first step towards this goal by inferring the geometric model from a depth image.

\textbf{Our approach.} We propose an approach to recover a 3D model of room layout and objects from an RGBD image. A major challenge is how to cope with the huge diversity of layouts and objects. Rather than restricting to a parametric model and a few detectable object classes, as in previous single-view reconstruction work, our models represent every layout surface and object with a 3D mesh that approximates the original depth image under projection. %The flexibility of our models is enabled through our approach (Fig.~\ref{fig:illustration}) to propose a large number of likely layout surfaces and objects and then compose a complete scene out of a subset of those proposals while accounting for occlusion, image appearance, depth, and layout consistency. 
We take a data-driven approach that proposes a set of potential object regions, matches each region to a similar region in training images, and transfers and aligns the associated labeled 3D models while encouraging their agreement with observations. During the matching step, we use CNNs to retrieve objects of similar class and shape and further incorporate support estimation to aid interpretation. We hypothesize, and confirm in experiments, that support height information will help most for interpreting occluded objects because the full extent of an occluded object can be inferred from support height.  The subset of proposed 3D objects and layouts that best represent the overall scene is then selected by our optimization method based on consistency with observed depth, coverage, and constraints on occupied space. The flexibility of our models is enabled through our approach (Fig.~\ref{fig:illustration}) to propose a large number of likely layout surfaces and objects and then compose a complete scene out of a subset of those proposals while accounting for occlusion, image appearance, depth, and layout consistency. 

\textbf{Detailed 3D labeling.} Our approach requires a dataset with labeled 3D shape for region matching, shape retrieval, and evaluation. We make use of the NYUd v2 dataset~\cite{silberman2012indoor} which consists of 1449 indoor scene RGBD images, with each image segmented and labeled with object instances and categories. Each segmented object also has a corresponding annotated 3D model, provided by Guo and Hoiem~\cite{guo2013iccv}. The 3D labeling provides ground truth 3D scene representation with layout surfaces as 3D planar regions, furniture as CAD exemplars, and other objects as coarser polygonal shapes. However, the polygonal shapes are too coarse to enable comparison of object shapes. Therefore, we extend the labeling by Guo and Hoiem with more detailed 3D annotations in the object scale. Annotations are labeled automatically and are adjusted manually as described in Sec.~\ref{SecModel}. We evaluate our method on our newly annotated groundtruth. We measure success according to accuracy of depth prediction of complete layout surfaces, voxel occupancy accuracy and semantic segmentation performance. 
%Experiments on object retrieval demonstrate better region classification and shape estimation compared with the state-of-the-art. Our performance of scene composition shows better semantic segmentation results and competitive $3$D estimation results compared with the state-of-the-art.

%For $3$D scene representation, we formulate it in a joint physical reasoning way, considering both local matching property like object shape, and global spatial coherency like 3D object exclusion, and whole scene object probability consistency. We apply a greedy with restart algorithm, to optimally select subset candidates as proper scene components.

%This paper is an extension of our previous work by Guo et al.~\cite{guo2015predicting} that predict full 3D scene parsing from an RGBD image. Compared to our previous work, we demonstrate better performance both quantitatively and qualitatively. We use CNNs to classify regions and retrieve objects with similar shapes, we incorporate support inference to aid region classification in images. What's more, we refine the NYUd v2 dataset with detailed 3D shape annotations for all the objects in each image for better validation. We conduct more
%extensive experiments and include more detailed explanation of our method.

Our main \textbf{contributions} are:
\begin{enumerate}
%\vspace{-0.5em}
    \item We refine the NYUd v2 dataset with detailed $3$D shape annotations for all the objects in each image. The labeling achieves a better representation of both ground truth object shape and whole scene depth. %We perform a semi-automatic approach and can be used to label for labelling other RGBD scene datasets.
    %\vspace{-0.5em}
    \item We propose an approach to recover a 3D model of room layout and objects from an RGBD image, arguably producing the most detailed and complete 3D scene parses to date. We take a data-driven approach, generating sets of potential object regions, matching to regions in training images, and transferring and aligning associated 3D models while encouraging fit to observations and overall consistency. We investigate impact of support estimation on scene parsing.
    %\vspace{-0.5em}
    %\item We demonstrate better performance in full $3$D scene parsing from single RGBD images compared with baseline method.
\end{enumerate}

In contrast to multiview 3D reconstruction methods, our approach recovers complete models from a limited viewpoint, attempting to use priors and recognition to infer occluded geometry, and parses the scene into individual objects and surfaces, instead of points, voxels, or contiguous meshes.  Thus, in some sense, we provide a bridge between the goals of interpretation from single-view and quantitative accuracy from multiview methods.

The overall technical contributions of this paper are substantial. Our approach tackles the challenging problem of recovering complete 3D model of indoor scene from single RGBD images. The proposed data-driven framework enables a detailed complete CAD model representation of 3D reconstruction~(Sec.~\ref{SecModel}) instead of sparse voxels, point clouds or surface meshes. To enable the feasibility of the complete CAD representation, as introduced in Sec.~\ref{sec:approach}, rather than performing retrieval on the entire image or superpixels, we use a more flexible region-to-region transfer to retrieve 3D shapes from a database. We then resolve conflicts in a final compositing process that accounts for fidelity to observed RGBD image. Extensive experiments demonstrate the success of our framework in handling complete 3D scene parsing quantitatively and qualitatively as shown in Sec.~\ref{exp:overall}. Our layout estimation outperforms existing approaches. We apply ablation study to investigate the effectiveness of the design choices in each of the main step in our framework in Sec.~\ref{subsec:evalretrieval}. We demonstrate improvements due to better region proposals. Moreover, our results from automatic region proposals are nearly as good as the result from ground truth regions, demonstrating the effectiveness of our scene composition and, more generally, finding a scene hypothesis that is consistent in semantics and geometry. Our CNN-based shape retrieval approach performs quantitatively better than the CCA based retrieval method by Guo et al.~\cite{guo2015predicting}. Furthermore, we study in-depth to verify our hypothesis that estimating support height of objects can lead to better classification, especially for occluded objects.

This paper is an extension of our previous work~\cite{guo2015predicting} that predicts full 3D scene parsing from an RGBD image. Our main new contributions are the refinement of the NYUd v2 dataset with detailed 3D shape annotations, the use of CNNs to classify regions and retrieve object models with similar shapes to a region, and use of support inference to aid region classification. We also provide more detailed discussion and conduct more extensive experiments, demonstrating qualitative and quantitative improvement. 

\section{Related work}
The most related work that recovers complete models from scene is proposed by Zhang et al.~\cite{zhang2014panocontext}, where they predict 3D bounding boxes of the room and all major objects inside, together with their semantic categories from an RGB 360$^\circ$ full-view panorama. Different from Zhang et al.: we interpret whole-room $3$D context with detailed 3D shapes and layout planes; our input is single image, which has limited field of view; and we make use of depth information, which eases the difficulty of interpreting geometry of visible surfaces. Other related topics are as follows.

\textbf{Room layout} is often modeled as a 3D box (cuboid)~\cite{hedau2009iccv,flint2011iccv,schwing2012eccv,zhang2013iccv,zhang2014panocontext}.  A box provides a good approximation to many rooms, and it has few parameters so that accurate estimation is possible from single RGB images~\cite{hedau2009iccv,flint2011iccv,schwing2012eccv, dasgupta2016delay} or panoramas~\cite{zhang2014panocontext}.  Even when a depth image is available, a box layout is often used (e.g.,ïœ\cite{zhang2013iccv}) due to the difficulty of parameterizing and fitting more complex models.  %However, many rooms and hallways are not exactly cuboid-shaped.
Others, such as~\cite{delage2006cvpr,lee2009cvpr, mallya2015learning}, estimate a more detailed Manhattan-structured layout of perpendicular walls based on visible floor-wall-ceiling boundaries.
Methods also exist to recover axis-aligned, piecewise-planar models of interiors from large collections of images~\cite{furukawa2009iccv} or laser scans~\cite{xiao2012eccv}, benefiting from more complete scene information with fewer occlusions.  We model the walls, floor, and ceiling of a room with a collection of axis-aligned planes with cutouts for windows, doors, and gaps, as shown in Figure~\ref{fig:layoutProc}.  Thus, we achieve similar model complexity to other methods that require more complete 3D measurements.

%\textbf{Categorical object region proposals.} 

Our use of \textbf{region transfer}
%We use a mixture of detailed models (for furniture) and approximate 3D polygonal models (for smaller objects).  One distinctive aspect of our problem is that we want to infer an approximate but complete 3D  model for every object that is at least partially visible from a single viewpoint.  The diversity of objects motivates us to take a retrieval-based approach, matching observed regions to training images and transferring associated categories and 3D models from them.
is inspired by the SuperParsing method of Tighe and Lazebnik ~\cite{tighe2010eccv}, which transfers pixel labels from training images based on retrieval.  Similar ideas have also been used in other modalities: Karsch et al.~\cite{karsch2012eccv} transfer depth, Guo and Hoiem~\cite{guo2012eccv} transfer polygons of background regions, Yamaguchi et al.~\cite{yamaguchi2013iccv} transfer clothing items.  Exemplar-based 3D modeling is also employed by Satkin and Hebert~\cite{satkin2013iccv} to transfer 3D geometry and object labels from entire scenes. %DeepContext~\cite{zhang2016deepcontext} aligns the observed RGB-D image with a predefined 3D scene template then reasons about the refinement. 
Rather than retrieving based on entire images~\cite{li2015jointembedding,zhang2016deepcontext} (which is too constraining) or superpixels (which may not correspond to entire objects), we take an approach of proposing a bag of object-like regions and resolving conflicts in a final compositing process that accounts for fidelity to observed depth points, coverage, and consistency.  In this way, our approach also relates to work on segmentation~\cite{silberman2012indoor, gupta2013cvpr,dollar2013iccv} and parsing~\cite{ren2012cvpr,banica2013iccv} of RGBD images and generation of bags of object-like regions~\cite{carreira2012pami,endres2010eccv,manen2013prime}. %Silberman et al.~\cite{silberman2012eccv} use both image and depth cues to jointly segment the objects into $4$ categories and infer support relations. Gupta et al.~\cite{gupta2013cvpr} apply both generic and class-specific features to assign $40$-class region labels. Their following work encodes both RGB and depth descriptors (HHA)~\cite{gupta2014eccv}: height above ground, angle with gravity and the horizontal disparity into the CNNs structure for region classification. Long et al.~\cite{long2015fully} introduce a fully convolutional network structure for learning better features.
The difference between our method and the above region transfer work is that: (1) we transfer 3D object models, rather than semantic labels; (2) we make use of larger, possibly overlapping object proposals, versus smaller, disjoint superpixels; and (3) we learn a distance function to improve the retrieval. 

\textbf{3D objects} are also often modeled as 3D boxes when estimating from RGB images~\cite{hedau2010eccv,xiao12nips,zhao2013cvpr,lin2013,zhang2014panocontext} or RGB-D images~\cite{lin2013}. However cuboids do not provide good shape approximations to chairs, tables, sofas, and many other common objects.  Another approach is to fit CAD-like models to depicted objects. From RGB images, Lim et al.~\cite{lim2013iccv,lim2014fpm} find furniture instances, and Aubry et al.~\cite{aubry2014cvpr} recognize chairs using HOG-based part detectors. In RGB-D images, Song and Xiao~\cite{song2014sliding} search for chairs, beds, toilets, sofas, and tables by sliding 3D windows and enumerating all possible poses. Song et al. then extend the method to detect a larger variety of classes~\cite{Song_2016_CVPR} with CNN-based approach and interprets semantic voxel representations from single depth map~\cite{song2016ssc}. Gupta et al.~\cite{gupta2015aligning} fit shape models of $6$ classes with poses to improve object detection. Xiang et al.~\cite{xiang2016objectnet3d} align 3D shapes from 100 categories to 2D images, contributing a large scale database for 3D object recognition. 
%These detector-based approaches often work with fixed categories because of the amount of training data required.
%Our approach does not aim to categorize objects but to find an approximate shape for any object, which could be from a rare or nonexistent category in the training set.
%We take an exemplar-based approach, matching regions from the input image to the training images and transferring corresponding 3D models.
Different from the above methods, our approach finds a detailed shape for any object and layout in the scene from a single RGBD image. %We take an exemplar-based approach, applying region-to-region retrieval to transfer similar 3D shape from training region to query. We also incorporate a per-object 3D alignment procedure that reduces the need to match to objects with exactly the same scale and orientation. We train our approach on our detailed 3D annotations for indoor scene in the NYUd v2 dataset. 
Our approach is efficient and light-weight compared to ScanNet~\cite{dai2017scannet}, which utilizes RGB-D video dataset to annotate instance-level semantic segmentations, since our model operates on single images instead of videos. Moreover, our detailed shape prediction can be utilized as an refining tool for existing dataset with 3D bounding box ground truth. %Rather than restricting to a parametric model and a few detectable objects, we make use of CNNs features to categorize objects into a larger variety of $81$ classes to distinguish infrequent objects in the scenes. Though semantic segmentation is not the main purpose of our method, we can infer region labels by projecting the $3$D scene models to the $2$D images, which is rough in contours but informs reasonable object shapes in the scene.%Incorporating category-based 3D detection, such as~\cite{song2014eccv}, is a promising direction for future work.
%One potential improvement for future work is to incorporate category-based 3D detection, such as~\cite{song2014eccv}.

%\textbf{Other techniques} that have been developed for scene interpretation from RGB-D images do not provide 3D scene interpretations but could be used to improve or extend our approach. For example, Silberman et al.~\cite{silberman2012eccv} infer support labels for regions, and Gupta et al.~\cite{gupta2014eccv} segment images into objects and assign category labels.  %These methods do not provide 3D scene interpretations but could be used to extend our representation to categories and physical support or to improve our process for object proposals and retrieval.  We leave these improvements to future work and focus on developing and evaluating the key ideas of the approach.

%We see many interesting directions for \textbf{further improvement} on the baseline framework. 1) Improving proposals /segmentation and incorporating object categories.  2) Incorporating support relations.  %modeling object context such as chairs tend to be near/under tables, and modeling self-similarity such as that most chairs within one room will look similar to each other.

\begin{figure*}[t]
\begin{center}
\includegraphics[width=1.0\linewidth]{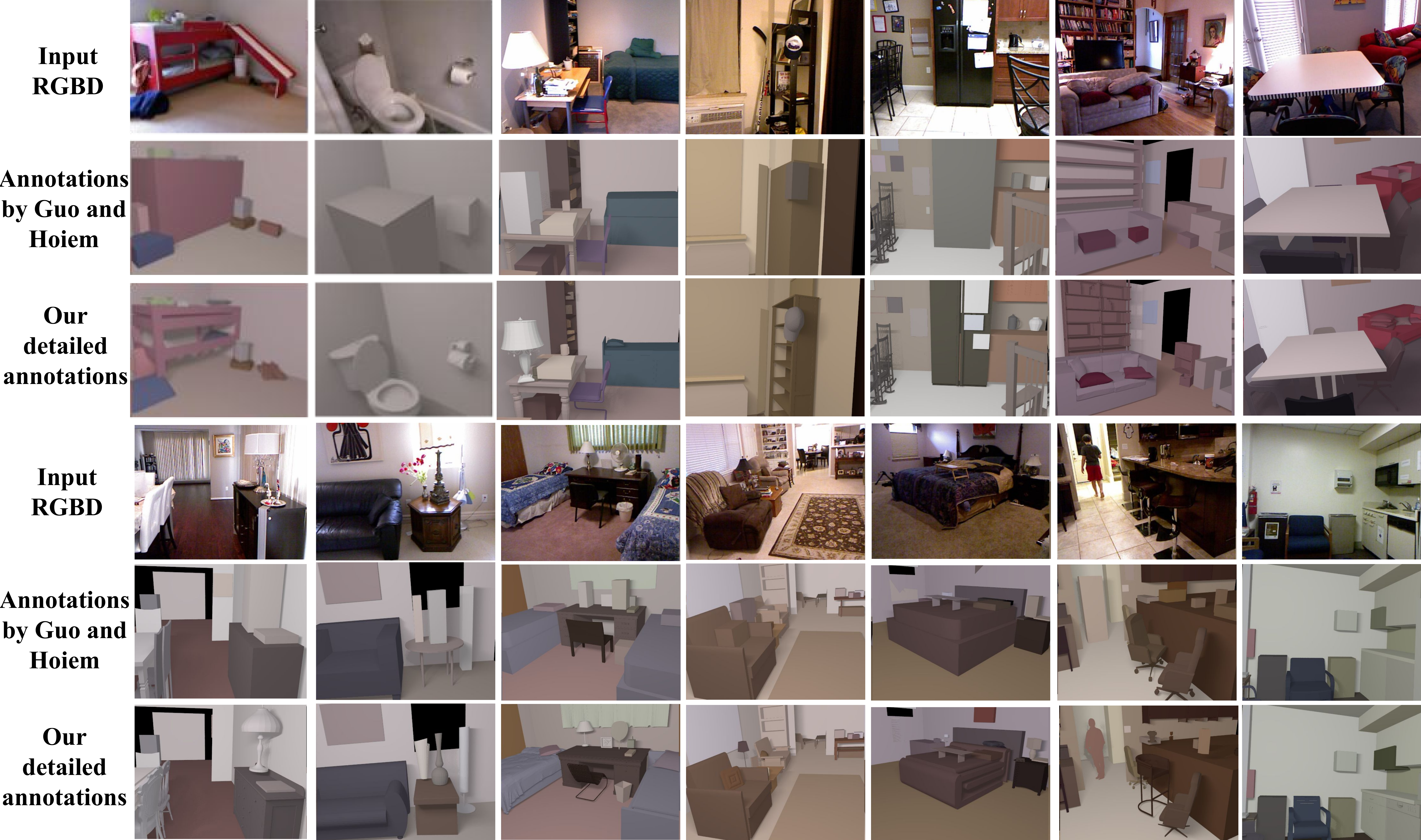}
\end{center}
   \vspace{-1em}
   \caption{\textbf{Samples of our detail $3$D annotation in NYUd v2 dataset}.We show from the left column to the right with different levels~(larger to smaller) of relative depth error improvements compared with Guo and Hoiem\cite{guo2013iccv}. Our annotations are much detailed in object shape scale.
   }
   \vspace{-1em}
\label{fig:dataset}
\end{figure*}
\begin{figure}
\begin{center}
%\fbox{\rule{0pt}{0.8in} \rule{0.5\linewidth}{0pt}}
%\vspace{-1em}
   \includegraphics[width=.9\linewidth]{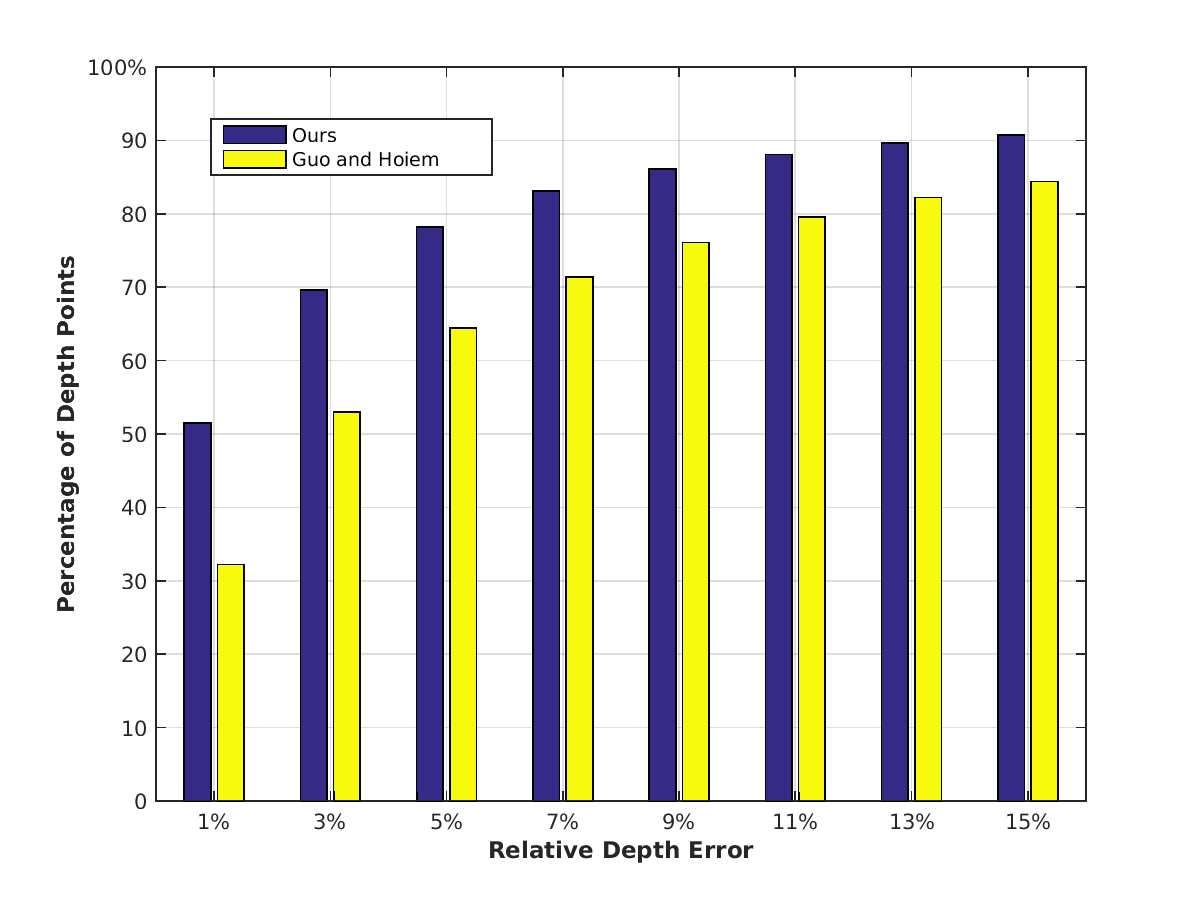}
\end{center}
\vspace{-1em}
   \caption{Cumulative relative depth error of our detailed $3$D annotations and the $3$D annotations by Guo and Hoiem~\cite{guo2013iccv} in NYUd v2 dataset.}
   \vspace{-1em}
\label{fig:dataset_d}
\end{figure}

\textbf{Support height estimation.} Guo and Hoiem~\cite{guo2013iccv} localize the height and full extent of support surfaces from one RGBD image. In addition, object height priors have shown to be crucial geometric cues for better object detection in both 2D~\cite{hoiem2008putting,walk2010disparity} and 3D~\cite{lin2013,song2014sliding,gupta2015aligning}. Deng et al.~\cite{deng2015semantic} apply height above ground to distinguish objects. We propose to use objects' support height to aid region class interpretation, which helps by distinguishing objects that appear at different height levels: e.g. chairs should be on the floor and alarm clocks should be on the table, and by inferring the full extent of an occluded object.

\section{Detailed $3$D annotations for indoor scenes}\label{SecModel}

%There's emerging RGBD indoor scene dataset with amodal $3$D labeling \cite{song2015sun,guo2013support}. 
We conduct our experiments on the NYUdv2 dataset~\cite{silberman2012indoor}, which provides complete 3D labeling of both objects and layouts of 1449 RGB-D indoor images. %with each image segmented and labeled into object instances and categories. 
Each object and layout has a 2D segment labeling and a corresponding annotated 3D model, provided by Guo and Hoiem~\cite{guo2013iccv}. The $3$D annotations use 30 models to represent 6 categories of furniture that are most common and use extruded polygons to label all other objects. These models provide a good approximation of object extent but are often poor representations of object shape, as shown in Fig~\ref{fig:dataset}. %Further, a detailed $3$D shape groundtruth of every object is also required to precisely evaluate complete $3$D scene representation. 
Therefore, we extend the NYUd v2 dataset by replacing the extruded polygons with CAD models collected from ShapeNet~\cite{shapenet2015} and ModelNet~\cite{wu20153d}. %We present a semi-automatic approach, based on rough $3$D annotation in \cite{guo2013support}.
%\subsection{Collecting CAD models for various classes}
%Each $3$D object has class label and corresponded 2D region marked in the image. 
To align name-space between datasets, we manually map all model class labels to the $633$-class $3$D object labels in NYUd v2 dataset. The shape retrieval and alignment process is performed automatically and then adjusted manually, as follows.

%\subsection{Semi-automatic labeling}
%We perform an auto-labeling method on all the 1449 images in the dataset.

\textbf{Coarse alignment.} For each ground truth $2$D region $r_i$ in the NYUd v2 dataset, we retrieve model set $M = \{M_i\}$ from our collected models that have the same class label as $r_i$. We also include the region's original coarse $3$D annotation by Guo and Hoiem~\cite{guo2013iccv} in the model set $M$, so that we can preserve the original labeling if no provided CAD models are better fit in depth. We initialize each $M_i$'s 3D location as the world coordinate center of the $3$D annotation labeled by Guo and Hoiem. We resize $M_i$ to have the same height as the $3$D annotation.

\textbf{Fine alignment.} 
Next, we align each retrieved 3D object model $M_i$ to fit the available depth map of the corresponding 2D region $r_i$ in the target scene. % five rotations -90 to 90, six scales; choose three with lowest depth error; then for each rotation and scale do ICP on translation
The initial alignment is often not in the correct scale and orientation; e.g., a region of a left-facing chair often resembles a right-facing chair and needs to be rotated. We found that using Iterative Closest Point to solve for all parameters did not yield good results.
%We search for scale and rotation in the ground plane and use Iterative Closest Point (ICP) to solve for translation. We also tried to use ICP to find the scaling and rotation in the ground plane, but that did not yield good results.
%Instead, we enumerate six scales and five rotations from -90 to 90 degrees, 
Instead, we enumerate 16 equally-spaced orientations from -180 to 180 from top-down view and allows 2 minor scale revision ratio as $\{1.0, 0.9\}$. We perform ICP to solve for translation initialized using scale and rotation, and pick the best ICP result based on the following cost function:
\begin{align}
\label{eq:deptherr}
&{\rm FittingCost}(M_i, T_i)=\notag\\
&C_{depth}\sum_{j\in r_i \cap s(M_i,T_i)} |\mathcal I_d(j)- \hat d(j;M_i,T_i)|\notag \\
&+\sum_{j\in r_i \cap \neg s(M_i,T_i)} C_{missing}\notag \\
&+C_{occ}\sum_{j\in \neg r_i \cap s(M_i,T_i)} \max(\hat d(j;M_i,T_i)-\mathcal I_d(j), 0)
\end{align}
where $T_i$ represents scale, rotation, and translation, $s(.)$ is the mask of the rendered aligned object, $\mathcal I_d(j)$ denotes the observed depth at pixel $j$ and $\hat d(j)$ means the rendered depth at $j$. The first term encourages depth similarity to the ground truth RGBD region. The second penalizes pixels in the proposed region that are not rendered. %($C_{missing}=0.3$ in our experiments); 
We loosely allow the rendered object depth in the scene to be further away than the sensor depth.  This is because depth map only gives the depth of the closest object, not all objects, due to possible occlusion. We do not want to penalize predicting a larger depth than is observed in the scene. The third term penalizes pixels in the rendered model that are closer than the observed depth image (so the model does not stick out into space known to be empty). %For efficiency, we evaluate the ten retrieved object models before alignment based on this cost function, discard seven, and align the remaining three by choosing $T_i$ to minimize ${\rm FittingCost}(M_i, T_i)$.

Based on the fitting cost of Eq.~\ref{eq:deptherr}, our algorithm picks the model $M_i$ with the best translation, orientation, and scale $T_i$. The fitting scales $T_i$ along different dimensions are the same. This helps simplify the search procedure for the alignment. We found that the large variety of CAD models in the ShapeNet dataset already provide enough shape candidates to select from, in regardless of the scales along the three dimensions of each shape.
%We define $s(\cdot)$ as the mask of the rendered $m_i$ in 2D, $I_d(j)$ as the observed depth of image $I$ at pixel $j$, $\hat{d}(j)$ as the rendered depth of $m_i$ at pixel $j$. We have:
%\begin{equation}\label{equ:fit}
%\begin{array}{rl}
%E_{fitting(m_i,t_i)}= &w_{m}\sum_{j\in r_i \cap \urcorner s(m_i, t_i)} 1
%+w_{d}\sum_{j\in r_i \cap s(m_i, t_i)} |\hat{d}(j|m_i, t_i) -I_d(j)| \\
%&+ w_{o}\sum_{j\in \urcorner r_i \cap s(m_i, t_i)} \max(\hat{d}(j|m_i, t_i) - I_d(j), 0) 
%\end{array}
%\end{equation}
We set the term weights $C_{depth}$, $C_{missing}$, $C_{occ}$ as $1.0, 0.9, 0.5$ using grid search in the validation set. The search criteria is to find the set of term weights that minimize both the rendered depth difference to the ground truth depth and the rendered 2D segmentation difference to the ground truth 2D region annotation. %following Gupta et al.~\cite{gupta2014eccv}. %where $w_{m}$ is $0.5$ for the term penalizing pixels in $r_i$ that are not rendered by $m_i$ in $2$D, $w_{d}$ is $1.0$ for the term encouraging depth similarity in $r_i$, $w_{o}$ is $0.9$ for the term penalizing pixels in the rendered model that are closer than the observed depth. 
For original 3D polygonal labeling, we %treat it the same as a candidate CAD shape model, 
fix $M_i$ and use equation 1 to search for the best fitting scale and translation that minimize the cost. For efficiency, we first obtain the top 5 models based on the fitting cost, each maximized only over the 16 initial orientations before ICP. For each of these models, we then solve for the best translation $T_i$ for each scale and rotation based on Eq.~\ref{eq:deptherr} and finally select the aligned model with the lowest fitting cost. %We obtain the best fit model $M_i$ with its best fit $T_i$ and $r_i$.

\textbf{Post-processing.} Automatic fitting may fail due to high occlusion or missing depth values. We manually conduct a post-processing check and refine bad-fitting models. Using a GUI, an annotator checks the automatically produced shape for each region.  If the result is not satisfactory, the annotator compares to other top model fits, and if none of those are good matches, then the fitting optimization based on Eq.~\ref{eq:deptherr} is applied to the original polygonal 3D labeling. This helps to ensure that our detailed shape annotations are a strict improvement over the original course annotations. %We in total use 3792 CAD models for labeling and only observe 10\% failures in the automatic fitting of CAD models.
In total, we fit 3792 different CAD models to the object regions and through manual checking observe failures in only 10\% of cases for automatic fitting. These failures are manually corrected.

\textbf{Validation}. Figure~\ref{fig:dataset_d} reports the cumulative relative error of the rendered depth of our detailed $3$D annotations compared with ground truth depth in NYUd v2 dataset. The relative error $r_D$ is computed as:

\begin{align}
\quad\quad\quad\quad\quad\quad r_D =\frac{1}{|S_I|} \sum_{I\in S_I}\sum_{p\in I}\frac{|d_p - \hat{d_p}|}{d_p}
\end{align}

where $S_I=\{I_1, I_2, \ldots, I_N\}$ is all the RGBD images in the dataset; $p$ represents a pixel in each image $I$; $d_p$ is the ground truth depth of pixel $p$ from sensor; and $\hat{d_p}$ is the rendered depth of the 3D label annotation at pixel $p$.
For comparison, we report the $r_D$ of the $3$D annotations by Guo and Hoiem~\cite{guo2013iccv}. Our annotations have more points with low relative depth error, %In table~\ref{tab:dataset_d} we evaluate the depth estimation by the metric of surface-to-surface distance~\cite{rock2015completing}, which is computed between 5000 sampled predicted depth points of each visible object shape and the ground truth object points. 
and achieve a better modeling of depth for each image.

%\begin{table}
%\begin{center}
%\caption{Depth estimation accuracy evaluated by surface distance.}
%\label{tab:dataset_d}
%\begin{tabular}{c|c|c}
%\toprule\noalign{\smallskip}
%& Guo and Hoiem~\cite{guo2013iccv} & Ours\\
%\midrule
%Surface distance (m) & 0.156 & 0.132
%\bottomrule
%\end{tabular}
%\end{center}
%\end{table}

\section{Approach}\label{sec:approach}

Given an RGBD image as input, we aim to find a set of layout and object models that fit RGB and depth observations and provide a likely explanation for the unobserved portion of the scene. A major challenge is how to cope with the huge diversity of layouts and objects. Existing approaches~\cite{Song_2016_CVPR,gupta2015aligning,song2014sliding} focuses on recovering a few detectable objects and are not feasible as an expressive scene representation to support whole scene interpretation. Song et al.~\cite{song2016ssc} propose a parametric model that employs a neural network to perform semantic completion from single depth images. However, a post-processing step is still needed for the approach by Song et al. to match candidate 3D shapes that fit both visible semantic information and scene regularity, e.g. no 3D overlap between shapes. Different from the above methods, we represent all layouts and objects in the scene as CAD models and propose a framework to parse both the visible and occluded portions of the layout and all observable objects, producing a complete 3D parse. Our resulting 3D parse is consistent with the observation and reasonable in 3D. We formulate this as:
\begin{align}
\{\mb M, \mb \theta\}=\arg \min (\rm{AppearanceCost}(\mathcal I_{RGBD},\mb M, \mb \theta)\nonumber\\
+\rm{DepthCost}(\mathcal I_D, \mb M, \mb \theta)+\rm{ModelCost}(\mb M, \mb \theta)).
\label{eq:global}
\end{align}
$\mathcal I_{RGBD}$ is the RGB-D image; $\mathcal I_D$ is the depth image alone; $\mb M$ is a set of candidate 3D layout surfaces and object models; and $\mb \theta$ is the set of parameters for each surface/object model, including translation, rotation, scaling, and whether each candidate model is included.  $\rm{AppearanceCost}$ encourages that object models should match underlying region appearance, rendered objects should cover pixels that look like objects (versus layout surfaces), and different objects should have evidence from different pixels.  $\rm{DepthCost}$ encourages similarity between the rendered scene and observed depth image.  $\rm{ModelCost}$ penalizes intersection of 3D object models.  %This optimization problem is hopelessly intractable: just choosing a subset of aligned models to fit the depth image is a hard combinatorial problem.

We propose to tackle this complex optimization problem in stages (Figure~\ref{fig:illustration}): (1) propose candidate layout surfaces and objects; (2) retrieve 3D models of the proposed candidate and improve the transformation of each surface/object model to the depth image; (3) choose a subset of models that best explains the scene.  Layout elements (wall, floor, and ceiling surfaces) are proposed by scanning for planes that match observed depth points and pixel labels and then finding boundaries and holes (Sec. ~\ref{subsec:layout}).
Parsing 3D objects is particularly difficult. We propose an exemplar-based approach, matching regions in the input RGBD image to regions in the training set, and transferring and aligning corresponding 3D models (Sec.~\ref{subsec:object}). We then choose a subset of objects and layout surfaces that minimizes the depth, appearance, and model costs using a specialized search (Sec.~ \ref{subsec:composition}). Despite the challenges of matching an optimization, we experimentally find that our approach produces results from automatic regions that are nearly as good as those produced from ground truth regions, even though using ground truth regions greatly simplifies the matching and selection process.

\subsection{RGBD image pre-processing}

We perform experiments on our newly re-annotated NYU v2 dataset, using the standard training/test split. The dataset consists of 1449 RGB-D images, with each image segmented and labeled into object instances and categories. Each segmented object has a corresponding annotated detailed 3D model labeled through the process in Sec.~\ref{SecModel}. Given a test image, we use the code from~\cite{silberman2012indoor} to obtain an oversegmentation with boundary likelihoods and the probability that each pixel $j$ corresponds to an object $P_{object}(j; I_{RGBD})$ (intsead of wall, floor, or ceiling). We use the code from~\cite{tighe2010eccv} to find the major orthogonal scene orientations, which is used to align the scene and obtain height value for each pixel.

\subsection{Layout proposals}
\label{subsec:layout}

\begin{figure}[t]
\begin{center}
\includegraphics[width=0.7\columnwidth]{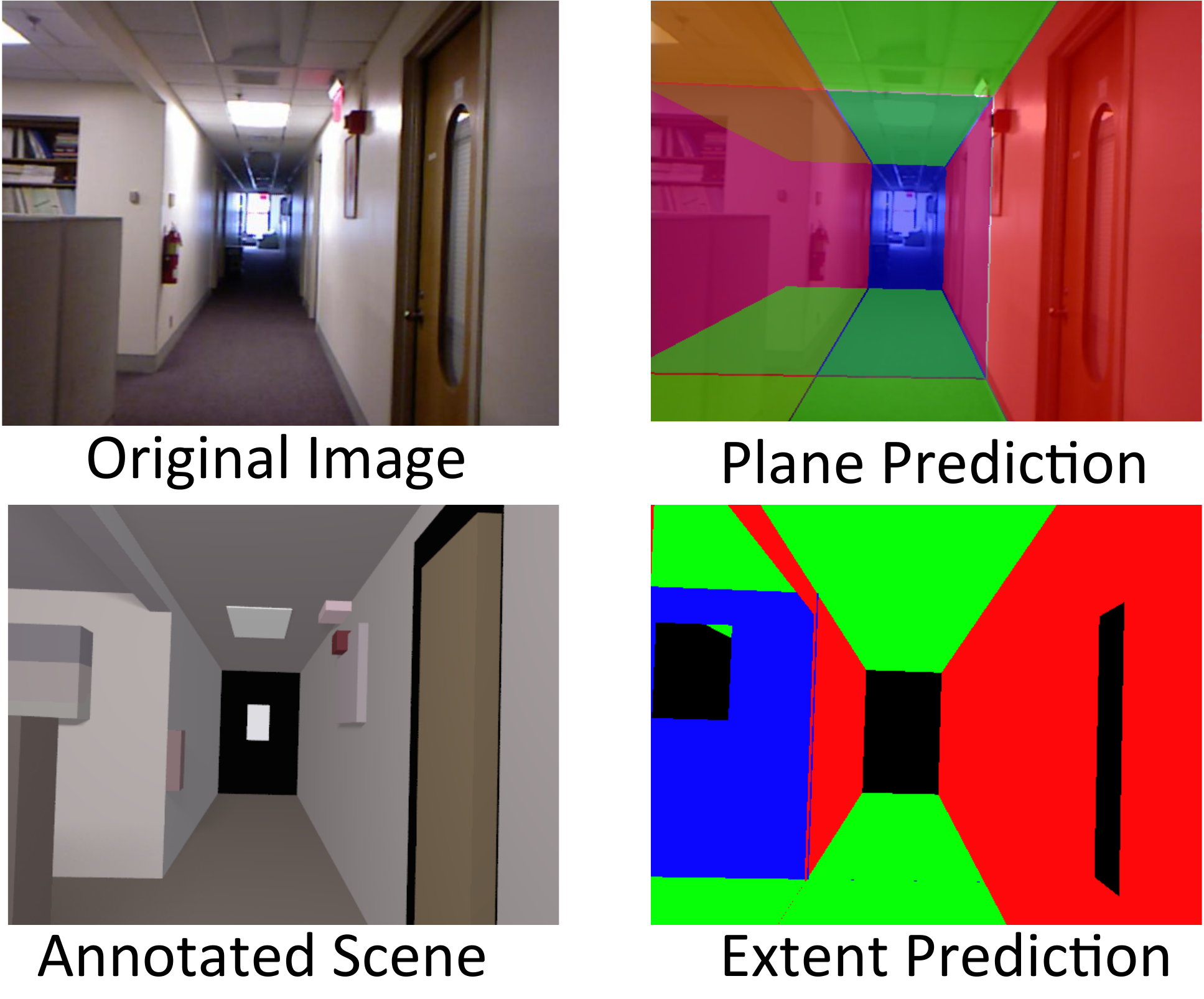}
\end{center}
\vspace{-1em}
\caption{\label{fig:layoutProc}
\textbf{Layout proposal}. We detect axis-aligned horizontal and vertical planes in the input depth image and estimate the extent of each surface.  Human-annotated layout is on the lower-left.}
\vspace{-1em}
\end{figure}

We generate layout proposals representing the full extent of possible layout surfaces, such as walls, floor, and ceiling. The layouts of these surfaces can be complex. For example, the scene in Figure~\ref{fig:layoutProc} has several ceiling-to-floor walls, one with a cutout for shelving, and a thin strip of wall below the ceiling on the left. The cabinet on the left could easily be mistaken for a wall surface.  Our approach is to propose a set of planes in the dominant room directions. These planes are labeled into ``floor'', ``ceiling'', ``left wall'', ``right wall'', or ``front wall'' based on their position and orientation. Then, the extent of the surface is determined based on observed depth points.

To find layout plane proposals, we aggregate appearance, depth, and location features computed from~\cite{silberman2012indoor} and train a separate linear SVM classifier for each of the five layout categories to detect planes. Appearance feature is the color histogram under YCbCr space. We define 
\begin{align}
{\rm p}(p_i ; P)=\mathcal N(dist(p_i, P), \sigma_p) \mathcal N(dist(n_i,P), \sigma_n)
\end{align} 
as the probability that a point with position $p_i$ and normal $n_i$ belongs to plane $P$, where the distances are point-to-plane and angular distance, and $\sigma_p=0.025$ and $\sigma_n=0.0799$ are based on Kinect measurement error.  Each pixel also has a probability of belonging to floor, wall, ceiling, or object, using code from~\cite{silberman2012indoor}.  The plane detection features are $f_1=\sum_i{{\rm p}(p_i ; P)}$; $f_2...f_5$, the sum in $f_1$ weighted by each of the four label probabilities; $f_6$, the number of points behind the plane by at least 3\% of plane depth; $f_7...f_{11}=(f_1...f_5)/f_6$; and $f_{12}$, a plane position prior estimated from training data.  Non-maximum suppression is used to remove weaker detections within 0.15m of a stronger detected plane.  Remaining planes are kept as proposals if their classification score is above a threshold, typically resulting in 4-8 layout surfaces.

The maximum extent of a proposed plane is determined by its intersection with other planes: for example, the floor cuts off wall planes at the base. Furthermore, hole cut-outs~(e.g. fireplace in a wall) are made by finding connected components of pixels with depth 5\% behind the plane, projecting those points onto the plane, fitting a bounding box to them, and removing the bounding box from the plane surface.  Intuitively, observed points behind the plane are evidence of an opening, and the use of a bounding box enforces a more regular surface that hypotheses openings behind partly occluded areas. Note that the fitted bounding box for hole cut-out on walls could be small compared to the ground truth due to occlusion. We observe that our simple approach produces excellent results in nearly all cases. Our method produces lower pixel labeling error for layout estimation than other approaches (see experiments in Sec.~\ref{exp:overall}).

\begin{figure*}[t]
\begin{center}

%\fbox{\rule{0pt}{1.2in} \rule{0.9\linewidth}{0pt}}
   \includegraphics[width=0.98\linewidth]{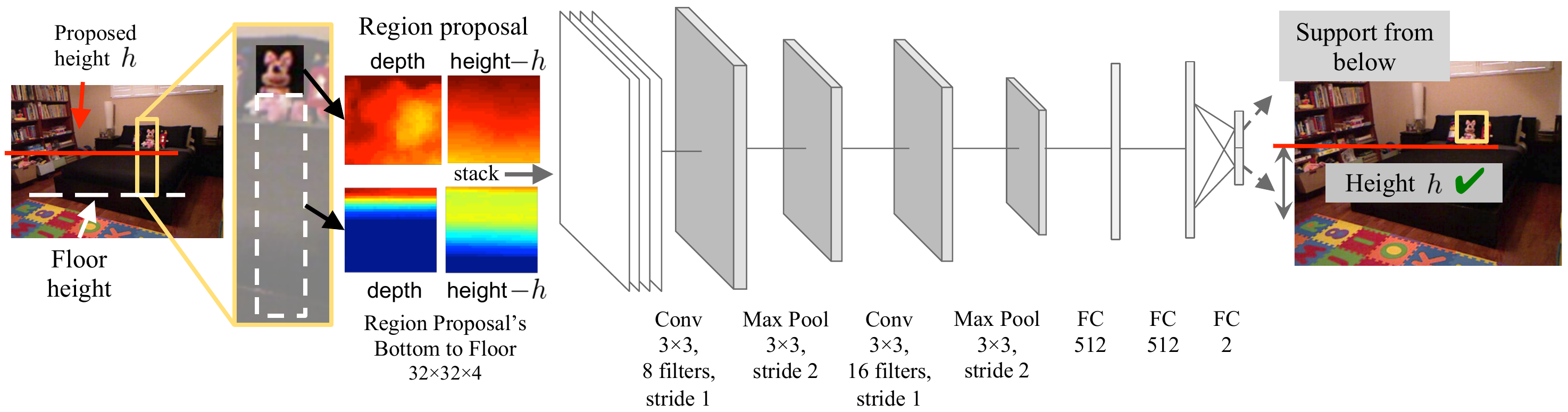}
\end{center}
\vspace{-1em}
   \caption{The CNN for predicting a candidate object's support height. We perform ReLU between the convolutional layer and the max pooling layer. Local response normalization is performed before the first fully connected (FC) layer. We add dropout with 0.5 before each FC layer during training.}
   \vspace{-1em}
\label{cnn:Height}
\end{figure*}

\subsection{Object candidates}
\label{subsec:object}
%We aim to produce a set of candidate regions with category label and aligned 3D shape from an RGBD image. %This is decomposed into region proposal generation and a 3D shape retrieval with category label and alignment for each region proposal. We describe the procedure in the following.
%\subsubsection{Region proposals}
%Our object region proposals involve a separate procedure from layout proposal.
To produce object candidate regions, we use the method for RGBD images by Gupta et al.~\cite{gupta2014learning} and extract top ranked $2000$ region proposals for each image. Our experiments show that this region retrieval is more effective than the method based on Prims algorithm~\cite{manen2013prime} used in our previous work~\cite{guo2015predicting}.   Likely object categories and 3D shapes are then assigned to each candidate region.

\subsubsection{CNN-based Shape Retrieval}\label{CNN_retrieval}
\begin{figure}
\begin{center}
%\vspace{-2em}
\includegraphics[width=0.95\linewidth]{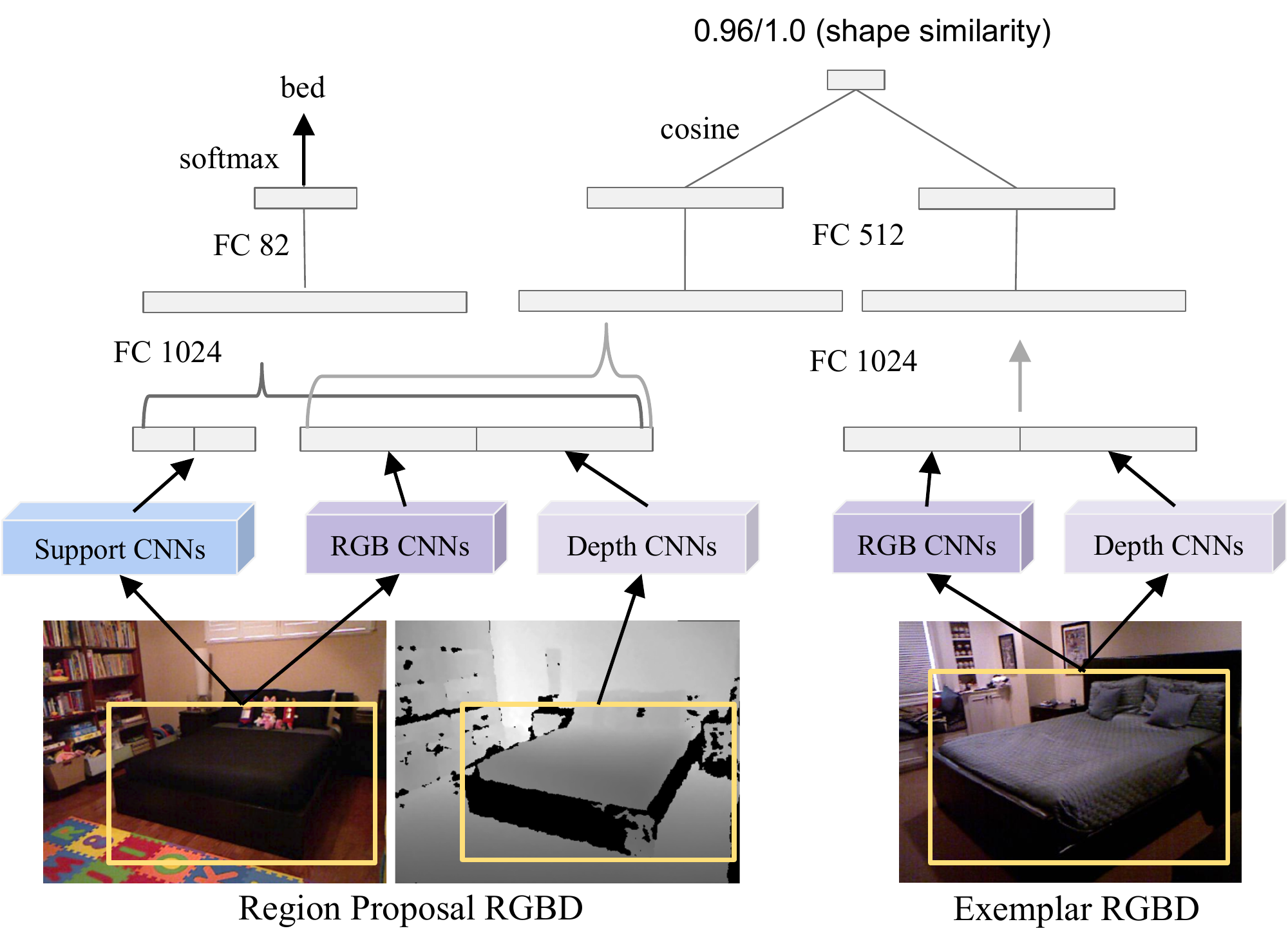}
\end{center}
\vspace{-1em}
   \caption{The CNNs for region classification (left) and similar shape retrieval (right). We perform ReLU and dropout with 0.5 after the first FC layer for both of the networks during training.}
   \vspace{-1em}
\label{fig:CNN_cls}
\end{figure}

%Shape similarity alone cannot guarantee valid retrieval: e.g., a whiteboard is similar to a door shape since they are both box-like and flat. Therefore, 
%We use object category probability and shape (appearance and orientation) similarity ranking as proxy for similarity when training the region matching function and transfer the annotated 3D object models.
%We consider the region-to-region retrieval based on both categorical characteristics and shape similarity. 
We train and use CNN networks to predict the object category and support height of each region, as shown in Fig.~\ref{cnn:Height} and Fig.~\ref{fig:CNN_cls}.  The support height is used as a feature for the object classification.  We also train and use a CNN with a Siamese network design to find the most similar 3D shape of a training object, based on region and depth features.  

\textbf{Support height prediction.} We predict support height for each object with the aim of better predicting class and position. As shown in Fig.~\ref{cnn:Height}, our support height prediction network jointly predicts the candidate support height probability and the support type: whether the object is supported from below or behind. We first find candidate support heights using the method of Guo and Hoiem~\cite{guo2013iccv} and use the network to estimate the probability of each candidate being the correct height. The support type prediction and the candidate support height probability prediction share the same convolution structure and the first two fully connected layers. Our network input consists two features: crops of  1) the depth maps and height maps of the region proposal and 2) the depth maps and height maps of the region that extends from the bottom of the region proposal to the estimated floor position. The former feature helps infer support type through its shape and the latter helps infer the support height through the vertical change in texture and depth. We found that having those two inputs together performs better for both suport height and support type predictions. To create the input feature vector, we subtract the candidate support height from the height cropped images, re-size all four crops to $32\times32$ patches, and concatenate channel-wise.  %with structure: $C(3,8,1) - RL - P(3,2) - N - C(3,16,1) - RL - P(3,2) - N - FC(512) - RL - D(0.5) - FC(512) - RL - D(0.5) - FC(2)$.

In the test set, we identify the closest candidate support height with $92\%$ accuracy, with an average distance error of $0.18$m. As a feature for classification, we use the support height relative to the camera height, which leads to slightly better performance than using support height relative to the estimated ground. This is because dataset images are taken from consistent heights but estimated ground height may be mistaken. 

\textbf{Categorization.} Our classification network gets input of the region proposal's support height and type, along with CNN features from both RGB and depth. The network predicts the probability for each class as shown in Fig.~\ref{fig:CNN_cls}. To model the various classes of shapes in indoor scene, we classify the regions into the $78$ most common classes, % to relief inter-class shape ambiguity. 
which have at least 10 training samples. Less common objects are classified as ``other prop", ``other furniture" and ``other structure" based on rules by Silberman et al.~\cite{silberman2012indoor}. In addition, we identify a region proposal that is not representative for an object shape (e.g. a piece of chair leg region when the whole chair region is visible) as a ``bad region" class. This leads to our $78+3+1 = 82$-class classifier. The input support height and type are directly predicted by our support height prediction network. To create our classification features, we copy the two predicted support values 100 times each (a useful trick to reduce sensitivity to local optima for important low-dimensional features) and concatenate them to the region proposal's RGB and HHA features from Gupta et al.~\cite{gupta2014learning} in both the $2$D bounding box and masked region as in~\cite{hariharan2014simultaneous}. Experiments show that using the predicted support type and the support height improves the classification accuracy by about $1\%$, with larger improvement for occluded objects.

%\subsubsection{Predicting region's $3$D shape}
\textbf{Shape candidate retrieval.}  Using a Siamese network (Fig.~\ref{fig:CNN_cls}), we learn a region-to-region similarity measure that predicts 3D shape similarity of the corresponding objects. The network embeds the RGB and HHA features used in our classification network into a space where cosine distance correlates to shape similarity, as in~\cite{yih2011learning}. In training, we use surface-to-surface distance~\cite{rock2015completing}, the normalized point cloud distance between the densely sampled points on each of the 3D meshes, as the ground truth similarity. The surface-to-surface distance is calculated from two centered complete 3D shapes complete 3D meshes, and thus no self-occlusion shall be considered during ground-truth computation. We train the network to penalize errors in shape similarity orderings. Each region pair's shape similarity score is compared with the next pair's among the randomly sampled batch in the current epoch and penalized only if the ordering disagrees with the ground truth similarity. We attempted sharing embedding weights with the classification network but observed a $1\%$ drop in classification performance.  We also found predicted class probability to be unhelpful for predicting shape similarity.

\textbf{Candidate region selection.} We apply the above retrieval scheme to each of the $2000$ region proposals in each image, obtaining shape similarity rank compared with all the training samples and $81$-object class and non-object class probability for each region proposal. In order to reduce the number of retrieved candidates before the scene interpretation in Sec.~\ref{subsec:composition}, we first reduce the number of region proposals using non-maximal suppression based on non-object class probability and threshold on the non-object class probability. We set the threshold to obtain $190$ region proposals for each image, on average. We select the two most probable classes for each remaining region proposal and select five most similar shapes for each class, leading to ten shape candidates for each region proposal. 

Then, we further refine these ten retrieved shapes. We align each shape candidate to the target scene by translating the model using the offset between the depth point mass centers of the region proposal and the retrieved region. %The height $h$ of object is constrained by distance from highest point in query region to the estimated support height. Since the highest point might be occluded, we check if $h$ conforms to the height distributions $N(\mu_c, \sigma_c ^2)$ for class $c$; if not, we set $h=\mu_c$. $\mu_c$ and $\sigma_c$ are learned from train set. 
We then perform ICP and use a similar method to that described in Section.~\ref{SecModel} to speed up the process. We use a grid of initial values for rotation and scale and pick the best one for each shape based on the fitting energy in Eq.~\ref{eq:deptherr}. We set the term weight: $C_{missing}$ as $0.6$, $C_{depth}$ as $1.0$, $C_{occ}$ as $0.9$ based on a grid search on the validation set. %we select top $3$ possible classes, retrieve the top $3$ most similar groundtruth meshes for each class. of 8 equally spaced interval from $-\frac{\pi}{2}$ to $\frac{\pi}{2}$, scale $[1.2, 1.12, 1.06, 1, 0.94, 0.88, 0.83]$, 
The overall selection costs on average 3.7 seconds for each object. We tried using the estimated support height for each region proposal for aligning the related 3D shape models but observed a worse performance in the scene composition result. This is because a relatively small error in object's support height estimation can cause a larger error in fitting. 

Finally, we select the two most promising shape candidates based on the following energy function,
\begin{align}
E_l(m_i) =& w_f E_{fitting}(m_i, t_i)\notag\\
&+ w_c \log P(c_i|r_i) + w_b \log P(b_i|r_i)
\end{align}
$E_{fitting}$ is the fitting energy defined in Eq.~\ref{eq:deptherr} that we used for alignment. $P(c_i|r_i)$ and $P(b_i|r_i)$ are the softmax class probability and the non-object class probability output by our classification network for the regon proposal $r_i$. We normalize $P(c_i|r_i)$ to sum to $1$, in order not to penalize the non-object class twice in the energy function. We set the term weights $w_f=1.0$, $w_c=-1500$, $w_b=1300$ using a grid search. Note that $E_{fitting}$ is on the scale of the number of pixels in the region.

\textbf{Training details.} We first find meta-parameters on the validation set after training classifiers on the training set.  Then, we retrain on both training and validation in order to report results on the test set. We train our networks with the region proposals that have the highest (and at least 0.5) $2$D intersection-over-union (IoU) with each ground truth region in the train set. We train the support height prediction network with the ground truth support type for the region proposal and set the ground truth support height as the closest support height candidate that is within $0.15$ meters from the related $3$D annotation's bottom. For training regions that are supported from behind, we do not penalize the support height estimation, since our support height candidates are for vertical support. For training the classification network, we also include the non-object class region proposals that have $<0.3$ IoU with the ground truth regions. We randomly sample the same number of the non-object regions as the total number of the object regions during training. To avoid unbalanced weights for different classes, we sample from the dataset the same number of training regions for each class in each epoch. When training the shape similarity network, we use the ground truth detailed CAD annotation obtained from Sec.~\ref{SecModel}. When conducting the shape candidate retrieval, the pool of 3D CAD models is the detailed annotation from the training split of the NYUd v2 dataset. We translate each 3D model to origin and re-size to $200\times200\times200$-voxel cuboid before computing the surface-to-surface distance.
%, which yields better performance.
%\textbf{Testing.} For each test region proposal, we predict its similarity rank with all $2$D regions in the train set. Note that the embedding of training regions could be precomputed only once.
We use ADAM~\cite{kingma2014adam} to train each network with the hyper-parameter of $\beta_1 = 0.9$, $\beta_2 = 0.999$. The learning rate for each network is: support height prediction $0.0008$, classification $0.003$ and Siamese network $0.0001$.

\subsection{Scene composition}
\label{subsec:composition}
Finally, given a set of candidate objects from Sec.~\ref{subsec:object} and layout plane candidates from Sec.~\ref{subsec:layout}, we need to choose a subset that closely reproduces the original depth image when rendered, adhere to pixel predictions of object occupancy, and correspond to minimally overlapping 2D regions and 3D aligned models. Instead of fitting each model to the depth map separately, we integrate the object and layout proposals together, optimize on proposals and compose a complete scene. The models $M=\{M_i\}$ and their alignment $T=\{T_i\}$ are fixed. The remaining parameters are $\mb y$ with $y_i\in\{0,1\}$ indicating whether each layout or object model is part of the scene. We choose $\mb y$ to minimize Equation~\ref{eq:comp}:
%\small
\begin{align}
\label{eq:comp}
&{\rm selectionCost}(\mb y)= \notag\\
&\sum_j {\rm clip}(|\log_2\frac{\hat d(j;\mb M, \mb T,\mb y)}{\mathcal I_d(j)}|-\log_2(1.03), [0~~1]) ~~+ \notag \\
&\sum_j |{\rm isObject}(j;\mb M,\mb T,\mb y)- {\rm P_{object}}(j; \mathcal I_{RGBD})| ~~+  \notag \\
&\sum_j \max(\sum_{i} y_i r_i(j)-1,0) ~~+ \notag \\
&\sum_{i,k>i} y_{i} y_{k} {\rm overlap3d}(M_{i},T_{i},M_{k},T_{k}).
\end{align}

%\normalsize
Here $j$ represents the rendered pixel for each candidate shape. $\hat d$ renders object models selected by $\mb y$. The first term minimizes error between rendered model and observed depth. We use log space so that the error matters more for close object.  We subtract $\log_2(1.03)$ and clip at 0 to 1 because errors less than 3\% of depth are within noise range, and we want to improve only reduction of depth if the predicted and observed depth is within a factor of 2.
%, ignoring errors larger than 3\% of depth because they are within noise range.  We clip at 1, so that reducing depth error is not rewarded unless the model depth is within a factor of 2 of observed depth.
The second term encourages rendered object and layout pixels to match the probability of object map ($P_{object}$) of each image based on~\cite{silberman2012indoor}. $r_i(j)$ is whether pixel $j$ belongs to the model $M_i$ rendered in 2D. The function $\rm isObject(j)=1$ identifies that a pixel corresponds to an object, rather than a layout model. The third term penalizes choosing object models that correspond to overlapping region proposals (each selected object should have evidence from different pixels).  The second and third term, combined with region retrieval, account for the appearance cost between the scene model and observations. The fourth term penalizes 3D overlap of pairs of aligned models, to encourage scene consistency.  To improve efficiency of this terms' computation, we approximate the object occupancy by rendering the closest and furthest points of the object at each pixel. We assume that all pixels between are occupied, and penalize using 3D overlap -- the length of the range of overlapping occupied depth, summed over each pixel. 

The depth rendering and consistency terms of Equation~\ref{eq:comp} involve high-order dependencies, leading to a hard binary problem. %The information captured in Equation~\ref{eq:comp} encourages depth similarity, compatibility to object and penalizes overlap between 2D rendered regions and 3D shape. Thus, we must solve a hard binary program in which some terms depend jointly on multiple entries in $\mb y$ due to occlusion.
We initialize $\mb y= \mb 0$ and estimate the solution in three steps. First, we perform greedy search to minimize depth error (minimize ${ \rm selectionCost}$ while weighting first term by a factor of 10) by iteratively adding the next candidate that maximizes ${ \rm selectionCost}$ gain. Then, this solution is used to initialize a hill-climbing search on ${\rm selectionCost}$ (experiments indicated insignificant benefit to weighting terms). The hill-climbing search iteratively adds ($y_i=1$) or removes ($y_i=0$), where $i$ indicates the candidate model that results in the greatest cost reduction, until no change yields further improvement. Finally, for all layout proposals and a subset of object proposals that are not yet selected, our algorithm tries adding the proposed model and removing all models whose renderings overlap and keeps the change if the cost is reduced.  %The subset is the set of object models that are based on regions that were already selected or that were the second best choice for addition at any step of the greedy search.  %A simple greedy optimization is not sufficient to yield good results due to complicated interactions of models with occlusion, and
In experiments, we found this search procedure to outperform a variety of other attempted methods including general integer programming algorithms and relaxations.

\section{Experiments}

%\subsection{Evaluation of scene composition}
The Overall performance is evaluated on the final complete 3D scene prediction for objects and layouts using the ground truth detailed 3D annotation (in Sec.~\ref{exp:overall}). We compare our current approach to our previous approach (Guo et al.~\cite{guo2015predicting}).  For some measures, we also compute the performance of the detailed 3D ground truth annotations as an upper bound. We investigate the effectiveness of the design choices in each of the main steps: region proposal, classification and shape retrieval, and scene composition. We also investigate the effects of incorporating support inference. To avoid overfitting to the NYUv2 dataset, we train our approach on the standard training split, tune all the parameters on the validation split and report results on the test split.

\textbf{Improving on Guo et al.~\cite{guo2015predicting}.} Our current approach improves on our previous work (\textbf{``Guo et al.''}) in two aspects.  First, we use region proposals from Gupta et al.~\cite{gupta2014learning} (``MCG'') instead of the region proposals generated by randomized Primâs algorithm~\cite{manen2013prime}. In Guo et al.'s method, region proposals are generated by first starting with an oversegmentation and boundary strengths by Silberman et al.~\cite{silberman2012indoor}. A neighborhood graph is then created, with superpixels as nodes and boundary strength as the weight connecting adjacent superpixels. Finally, the randomized Primâs algorithm is applied on the graph to obtain a set of region proposals. The seed region of the Primâs algorithm is sampled according to the objectness of the segment, so that the segments that are confident layout are never sampled as seeds. Size constraints and merging threshold are used to produce a more diverse set of segmentations. Regions that are near-duplicates of other regions are suppressed to make the set of proposals more compact. For each image, 100 candidate region proposals are generated. Second, our method considers both object category and shape similarity for retrieval, while Guo et al. use only object category as a proxy for object similarity. Guo et al. apply canonical-correlation analysis (CCA) to find embeddings of visual features that improve retrieval~\cite{gong2013ijcv}. CCA finds pairs of linear projections of the two views $\*\alpha^T\*X$ and $\*\beta^T\*Z$ that are maximally correlated. Here $\*X=\{\*x_i\}$ are the feature vectors of the regions. In experiments, Guo et al. use the combination of 3D features and RGBD-SIFT, which were found to provide better performance than pre-trained CNN features. $\*Z=\{\*z_i\}$ are the label matrix, which contains the one-hot indicator vectors of the object category labels of the corresponding region. The similarity weight matrix is then computed as $\*W=\*\alpha\*\alpha^T$, which is used to parametrize the distance metric:
\begin{align}
\quad\quad\quad\text{dist}_W(\*x_i, \*x_j)=\sqrt{(\*x_i-\*x_j)^T\*W(\*x_i-\*x_j)}
\end{align}
In the experiments for Guo et al., the top 3 object models nearest to each region proposal according to $\text{dist}_W$ are retrieved. 

\textbf{Detailed ground truth labeling as an upper bound.} We include the evaluation of our detailed ground truth annotation for the pixel-wise measures in Sec.~\ref{SecModel}, denoted by \textbf{``3D ground truth''}. Note that even the ``3D ground truth'' sometimes does not achieve high accuracy, because rendered models may not follow image boundaries and some small objects are not modeled in annotations.  

\textbf{Comparison with Deep Sliding Shapes~\cite{Song_2016_CVPR}~(``DSS'').} We compare our approach with %the scene parsing method 
DSS, the state-of-the-art amodal 3D object detector in RGBD images. Given single RGBD image, DSS %aims at 3D object detection and 
predicts %on single RGBD image with 
multiple~(around 300 after NMS) overlapping tight 3D bounding boxes and a confidence score for each object class. Each object class is predicted independently, therefore there is no constraint on 3D overlap between them. The evaluation criteria is the mean average precision~(mAP) for each class. %a large number of 3D object proposal for the defined 19 common furniture classes, and use a separate network to predicts a softmax objectness score for each 3D proposal. 
Different from DSS, our approach aims at producing an exact reconstruction of scene composed by all coherent objects and layouts with detailed CAD representation, while penalizing on object 3D overlap to ensure consistency. We evaluate on final scene composition performance to match observed RGB and depth. Though our approach have different goal, metrics and optimization strategy compared with DSS, we can still convert the results to each other's format and compare. 

\textbf{Comparison with Semantic Scene Completion~\cite{song2016ssc}~(``SSCNet'').} We compare our approach with the scene-level approach SSCNet, which is the state-of-the-art semantic scene completion method from single depth image. SSCNet predicts occupancy in a voxel space of $60\times36\times60$ within the view frustum of the scene. Each predicted voxel is assigned to either one of the 3 layout labels~(wall, ceiling, floor) or one of the 8 common furniture classes (e.g. bed, chair, table). Both SSCNet and our approach predict semantic labeling of the observed scene. Different from SSCNet, our framework %instead of applying parametric modeling of scenes, performs a data-driven approach and 
utilizes RGB information in addition to observed depth. Our approach finds a mesh-based representation for any object and layout in the scene, compared with SSCNet which produces a limited number of classes under the voxel-based representation and is not flexible in rendering the scene with a higher resolution. For comparison, we align and voxelize our predicted 3D scene model to the same configurations as SSCNet and evaluate on their metric.

\subsection{Performance measures}

%We show the quantitative evaluation of our two improvement on Guo et al.'s method in table~\ref{table:ablation}. 
We evaluate our 3D scene completion approach on our newly annotated NYUd v2 dataset. %We report the result our CNNs based approach on our new annotations. %Although both are our approaches, we denote the latter one as Guo et al. in the following to make a distinction.
We use both 2D and 3D quantitative measures to evaluate different aspects of our solutions: (1) instance segmentation and semantic segmentation performance induced by rendering; %(2) categorization and shape estimation of retrieved regions; 
(2) label/depth accuracy of predicted layout surfaces; (3) voxel accuracy of occupied space and freespace. Among these, the depth accuracy of predicted layout surfaces and voxel accuracy of object occupancy are the most direct evaluations of our 3D model accuracy. %The retrieval categorization serves to evaluate feature suitability and metric learning.
The instance segmentation and semantic segmentation indicates accuracy of object localization and classification. %and are useful for the in-depth study of the improvements of our method compared with Guo et al.'s. 
Where possible, we compare on these measures to prior work and sensible baselines. 

%To compare with other method besides our previous approach, we compare with Deep Sliding Shapes~\cite{Song_2016_CVPR}~(Sec.~\ref{sec:DSS}) on Amodal object detection and compare with Semantic Scene Completion~\cite{song2016ssc}~(Sec.~\ref{sec:SSC}) on semantic scene completion. We explain the performance measures as below.

For the comparison with DSS, we directly use DSS's 3D bounding box prediction on the 654 test images of the NYUd v2 dataset. We use our detailed annotated ground truth to compare on our metrics. Since DSS has no layout prediction, we only compare on the 19 classes that DSS concerns. We use the following three metrics for parsing objects: 3D voxel object occupancy, 2D semantic segmentation and instance segmentation. Note that DSS produces multiple overlapping predictions with confidence. % score, 
We first filter out detections with score lower than 0.5 in each image. %We keep the top scored detection and remove other overlapping boxes for each class in one image, then keep the top scored one in the remaining boxes and remove other overlapping boxes until no box left. 
We then apply NMS with an IoU threshold of 0 for each class separately to obtain non-overlapping top ranked 3D boxes. DSS has no 2D segmentation mask for each detection, we thus project the 3D box on the 2D image to get the predicted object segmentation and depth. We ignore the layouts and other object classes in both our 3D/2D predictions and the ground truth. On the other hand, to compare on DSS's amodal 3D object detection metric~\cite{song2014sliding}, we extract a tight 3D bounding box around each of our predicted detailed shape and set the confidence of each box to 1. %We also evaluate on the 3D amodal object detection metric which is used in the DSS approach.
We also compare with the result of Sliding Shapes by Song et al.~\cite{song2014sliding}. Sliding Shapes predicts detailed shapes, while the DSS approach only predicts 3D bounding boxes. Same as~\cite{song2014sliding} and~\cite{Song_2016_CVPR}, we evaluate on the ground truth of five main furniture on the test set which is the intersection of NYUd v2 and the Sliding Shapes test set. 

For the comparison with SSCNet, we evaluate on SSCNet's metric given the input of the kinect depth map and RGB observations on the test split of the NYUd v2 dataset. Since our predicted complete scene model is aligned with the camera coordinate, we first align our predictions to the world coordinate given groundtruth floor plane. We then voxelize the scene into $60\times36\times60$ voxel grids to obtain the same resolution as in SSCNet and then compare. We map our 84-class semantic labeling to the 11-class used in SSCNet. We then evaluate on two metrics~\cite{song2016ssc}: (1) scene completion with precision, recall and voxel-level IoU; (2) semantic scene completion with voxel-level IoU. Though our method is not optimized for voxel-based evaluation, we can still compare and evaluate to validate our approach on the scene-level 3D parsing performance. %SSCNet gets input of single depth map only.

\subsection{Evaluation of scene composition}\label{exp:overall}

\begin{figure*}
\begin{center}
%\fbox{\rule{0pt}{2.5in} \rule{0.9\linewidth}{0pt}}
   \includegraphics[width=0.97\linewidth]{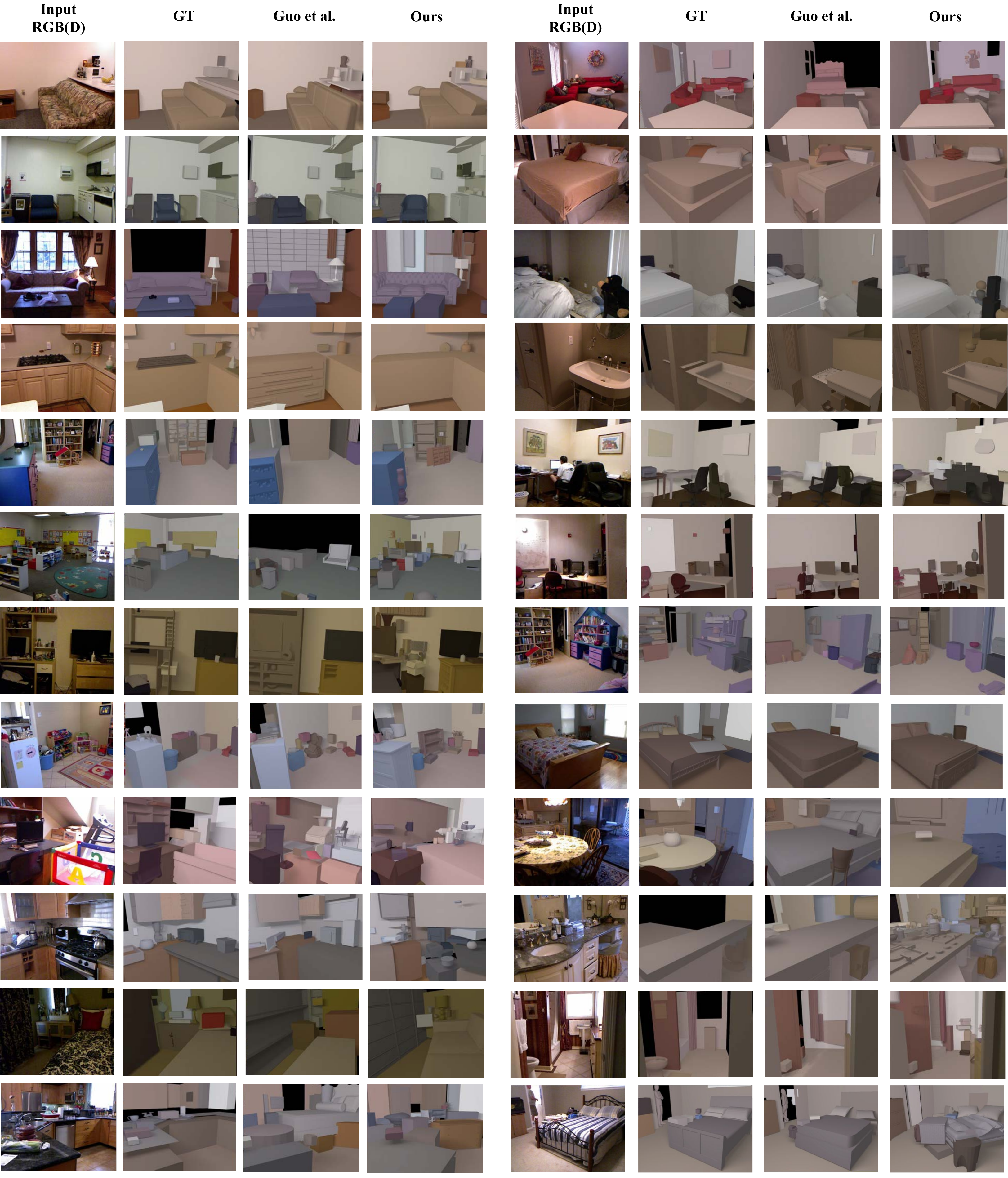}
\end{center}
\vspace{-1.5em}
   \caption{Qualitative results on scene composition with automatic region proposals. We randomly sample images from the top 25\% (first four rows), medium 50\% (row 6-9) and worst 25\% (last four rows) based on 84-class semantic segmentation accuracy.}
   \vspace{-1em}
\label{fig:final}
\end{figure*}

\begin{figure*}
\begin{center}
%\vspace{-1em}
%\fbox{\rule{0pt}{2.5in} \rule{0.9\linewidth}{0pt}}
   \includegraphics[width=0.97\linewidth]{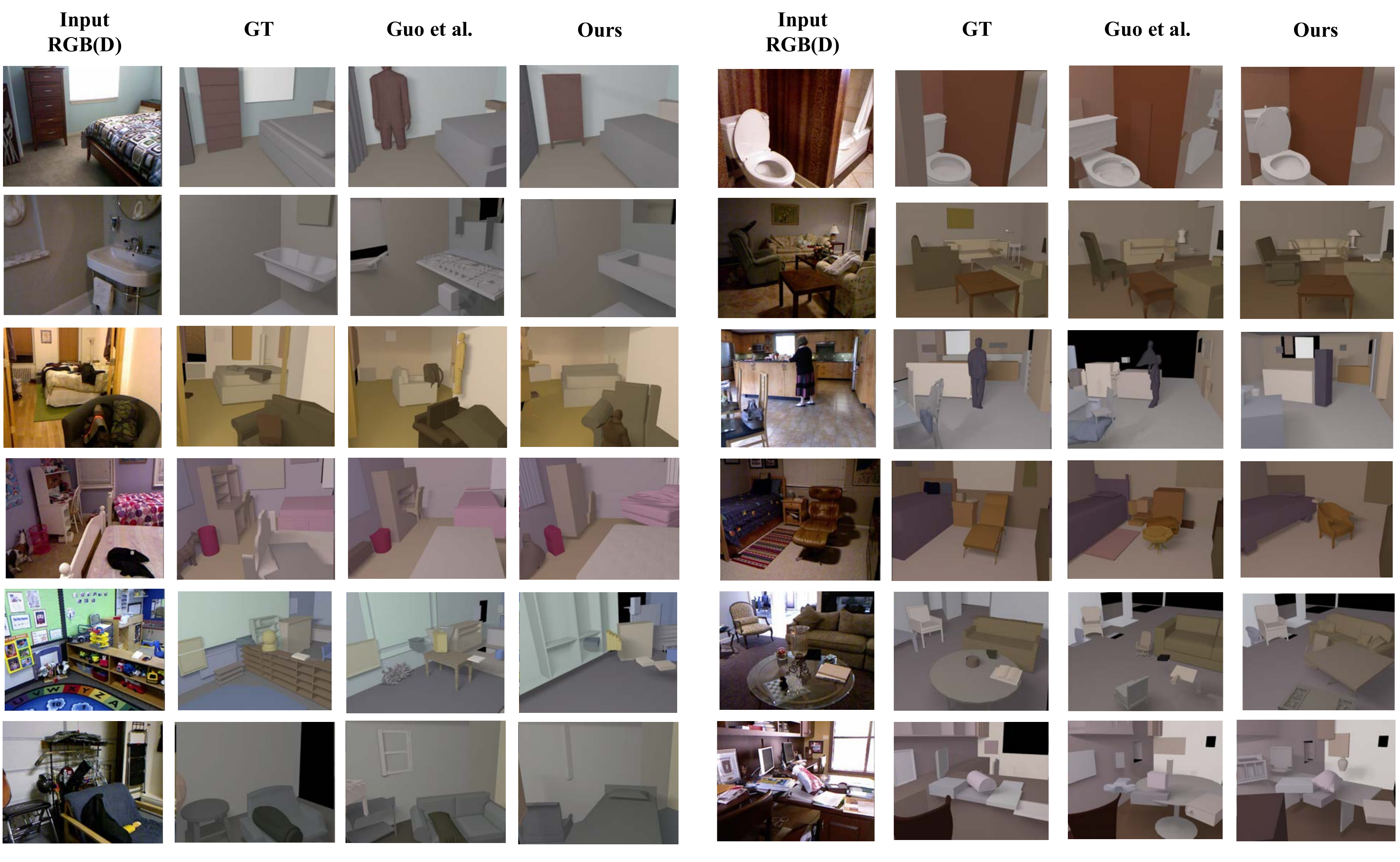}
\end{center}
\vspace{-1.5em}
   \caption{Qualitative results on scene composition given ground truth 2D labeling as region proposals. We randomly sample images from the top 25\% (first two rows), medium 50\% (row 3-4) and worst 25\% (last two rows) based on 84-class semantic segmentation accuracy.}
\label{fig:gt}
\end{figure*}

\textbf{Qualitative results.} We show representative examples of predicted 3D scenes in Figures~\ref{fig:final} and~\ref{fig:gt} of our approach, 3D ground truth, and Guo et al.'s method. %Sample qualitative results are shown in Fig.~\ref{fig:final} and Fig.~\ref{fig:gt}. 
Compared with the ground truth annotation, our method produces a similar layout parsing and a reasonable prediction and localization of main furniture in the scene. Compared with our previous approach of Guo et al., our method performs better, which results from the better region classification and shape estimation capability that will be analyzed in the ablation study Sec.~\ref{subsec:evalretrieval}. %Failure cases can be caused by bad pruning in region proposals and confusion between similar class.

\begin{figure*}
\begin{center}
%\vspace{-1em}
%\fbox{\rule{0pt}{2.5in} \rule{0.9\linewidth}{0pt}}
   \includegraphics[width=0.95\linewidth]{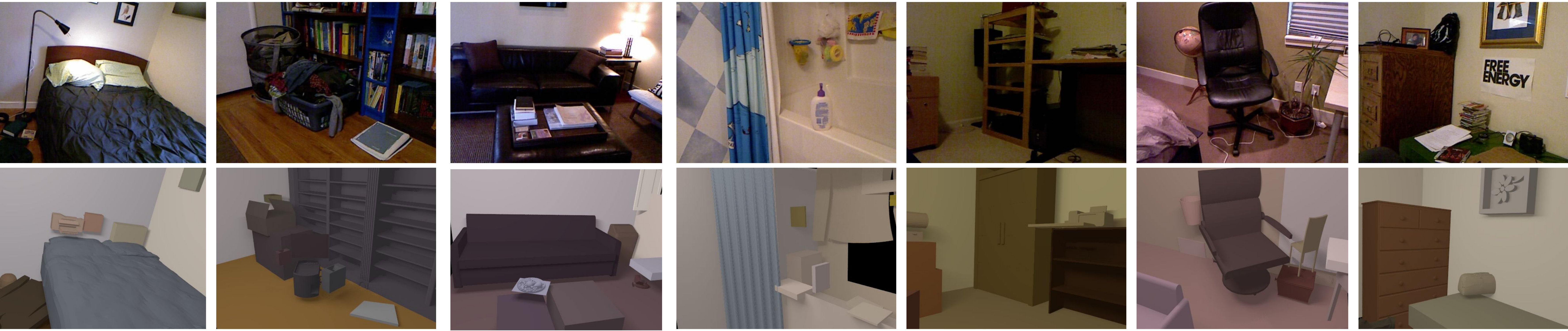}
\end{center}
\vspace{-1.5em}
   \caption{Qualitative results on SUN-RGBD dataset. Our method is able to estimate accurate 3D layout and the 3D occupancy of main furniture.}
\label{fig:sunrgbd}
\end{figure*}

To demonstrate the generalization capability to
other dataset, we show qualitative results on the SUN-RGBD dataset~\cite{song2015sun} in Figure~\ref{fig:sunrgbd}. Since SUN-RGBD dataset does not have detailed shape annotations to train on, we directly use our model trained on NYUd v2 dataset. Our method is able to obtain the accurate 3D parse of layouts and the main furniture like beds, shelves and desks. Errors mainly come from region proposal classification failure and missing small objects from the region proposal step, due to the domain difference between the two datasets.

\textbf{84-class instance segmentation.} We can infer region labels by projecting the $3$D scene models to the $2$D images. Though instance segmentation neglects the evaluation of 3D localization of shapes and layout, which is the main purpose of our method, it can be a reasonable evaluation of object localization and classification. We evaluate the 84-class instance segmentation (81 object classes and 3 layout classes: wall, ceiling, floor) on the inferred region labels of our scene composition result. The evaluation follows the protocol in RMRC~\cite{urtasun2013reconstruction} that computes the coverage of ground truth regions, weighted or unweighted by area, for each image.  The average coverage across all images is reported in  Table~\ref{tab:ins}.
%We compare our approach with the Guo et al's method. 
We see improvements, compared to Guo et al., due to both classification method (CNN vs. CCA) and region proposal method (MCG vs. Prim's). Also, it is interesting to see that using ground truth regions does not dramatically improve results, indicating that the region proposal and scene composition methods are highly effective. 

\begin{table}
%\vspace{-1em}
\begin{center}
\caption{Results of 84-class instance segmentation on both automatic region proposals and ground truth regions.}\label{tab:ins}
\resizebox{0.90\linewidth}{!}{
\begin{tabular}{c|cc}
\toprule\noalign{\smallskip}
\multirow {2}{*}{Method}  & Mean coverage & Mean coverage\\
& weighted & unweighted\\
\midrule
%\parbox[t]{2mm}{\multirow{5}{*}{\rotatebox[origin=c]{90}{Auto-region}}}& 
Guo et al.~\cite{guo2015predicting} & $48.12$ & $29.75$ \\
MCG+CCA & $50.43$&$32.91$\\
Ours w/o support & \textbf{50.43} & \textbf{34.34} \\
Ours& $50.23$&$33.92$\\
%\midrule
\midrule
Ours w/ GT support &$50.52$&$34.32$ \\
%\parbox[t]{2mm}{\multirow{4}{*}{\rotatebox[origin=c]{90}{GT-region}}}& 
%Guo et al.~\cite{guo2015predicting}& $51.77$ & $35.22$\\
%&Ours w/o support& $52.11$ & \textbf{35.64} \\
Ours w/ GT-region & 52.39 & $35.61$ \\
%&Ours w/ GT support& $52.53$ & $35.69$\\
\midrule
\midrule
 Detailed ground truth labeling & $65.42$ & $49.15$\\
\bottomrule
\end{tabular}}
\end{center}
\vspace{-1em}
\end{table}

\begin{table}
\vspace{-1em}
\begin{center}
\caption{Results of 84-class semantic segmentation on both automatic region proposals and ground truth regions.}
\label{table:sem}
\resizebox{0.90\linewidth}{!}{
\begin{tabular}{c|ccc}
\toprule\noalign{\smallskip}
%\multirow{2}{*}{} & 
\multirow{2}{*}{Method} & \multicolumn{3}{c}{Objects+Layouts}\\%\\ & \multicolumn{3}{c|}{Objects Only} & 
%\multicolumn{11}{c}{Sample Objects Result}\\
%\cline{3-19}
\cline{2-4}
   & avg class & avg instance & avg pixel\\% & avacc & fwavacc & pixacc & picture & chair & cabinet & pillow & bottle & books  & paper &table & box & window & door\\% & sofa & bag & lamp\\
\midrule
%\parbox[t]{2mm}{\multirow{4}{*}{\rotatebox[origin=c]{90}{Auto-region}}} & 
Guo et al.~\cite{guo2015predicting} & $6.96$ & $30.39$ & $43.85$\\% & &&& \\
 MCG + CCA & $8.04$ & $32.56$ & $46.17$ \\
Ours w/o support & $10.51$ & $32.45$ & $45.94$ \\%& &&& $14.76$ & $18.12$ & $20.82$ & $15.42$& $6.42$ & $5.63$ & $6.42$ & $18.45$ & $1.95$ & $14.90$ & $12.71$\\
Ours & \textbf{10.77} & \textbf{32.90} & \textbf{46.25} \\%& &&& $12.18$& $20.52$ &$24.78$ & $17.08$ & $5.92$ & $5.67$ & $9.57$ & $21.81$ & $4.04$ & $11.87$ & $10.28$\\
\midrule
Ours w/ GT support & $11.28$ & $33.71$ & $47.46$ \\%&  &&& $12.72$ & $24.94$ & $24.72$ & $17.43$ & $7.21$ & $6.01$ & $10.48$ & $20.04$ & $2.32$ & $13.44$ & $10.45$\\
%\midrule
%\parbox[t]{2mm}{\multirow{4}{*}{\rotatebox[origin=c]{90}{GT-region}}}&Guo et al.~\cite{guo2015predicting}& $10.35$ & $34.43$ & $51.39$ \\
%&Ours w/o support& $15.35$ & $37.09$ & $54.19$\\% &$13.52$ &$25.04$&$34.05$& $19.13$ & $34.98$ & $31.99$ & $24.86$ & $13.48$ & $4.43$ & $6.09$ & $29.72$ & $6.26$ & $6.82$ & $12.87$\\
Ours w/ GT-region& 16.63 & 37.61 & 54.86 \\%& $14.83$ & $25.90$& $35.11$& $18.95$ & $34.66$ & $37.79$ & $22.63$ & $11.08$&$4.54$ & $6.89$ & $27.41$ & $7.76$ & $6.96$ & $13.95$\\
%&Ours w/ GT support& $16.93$ & $38.11$ & $55.41$ \\%&$15.14$ &$26.60$& $35.85$& $19.44$ & $35.99$ & $38.45$ & $24.87$ & $10.71$&$4.22$ & $8.42$ & $30.28$ & $5.24$ & $6.76$ & $13.51$\\
\midrule
\midrule
 Detailed ground truth labeling & 67.85 & 77.53&85.78\\
\bottomrule
\end{tabular}
\vspace{-1em}
}
\end{center}
%\vspace{-2.5em}
\end{table}

\textbf{84-class semantic segmentation.} Semantic segmentation is evaluated by the percent of pixels that are correctly classified, either averaged across pixels (``avg pixel''), instances (``avg instance''), or classes (``avg class''), as reported in Table~\ref{table:sem}.  The average across pixels tends to be highest, since the most common and largest objects are more likely to be correctly labeled. Again, our approach outperforms Guo et al. due to improved region proposals and classification. Support prediction also helps here.  %Note that the lower results for MeanCovU is caused by the large diversity of classes with less frequency. We also include the evaluation of our detailed groundtruth annotation as the upper bound.
We also compute the $40$-class semantic segmentation to compared to the state-of-the-art~\cite{long2015fully}: $18.3$ average class / $33.8$ average instance / $47.7$ average pixel vs. their $34.0$/$49.5$/$65.4$.  Since our method projects 3D models of classified regions onto the images, it may not adhere to image boundaries as well as a direct semantic segmentation method and also may have more difficulty with small objects.

\begin{table}
\begin{center}
\caption{Depth error for visible and occluded portions of layouts.  Sensor error is the difference between the input depth image and annotated layout.}
\label{table:depth}
\resizebox{0.35\textwidth}{!}{
\begin{tabular}{c|ccc}
\toprule\noalign{\smallskip}
\multirow {2}{*}{Method}  & \multicolumn{3}{|c}{Layout Depth Error}\\
\cline{2-4}
 & overall & visible & occluded\\
\midrule
Sensor & 0.517 & \textbf{0.059}  & 0.739 \\
Guo et al.~\cite{guo2015predicting} & 0.166 & 0.074  & 0.204 \\

Ours& \textbf{0.150} & $0.074$ & \textbf{0.181} \\

\bottomrule
\end{tabular}
}
\vspace{-2em}
\end{center}
\end{table}

\begin{table}
\begin{center}
\caption{Pixel labeling error for layout surfaces with 5 categories (full dataset, 654 image).}
\label{table:layout}
\resizebox{0.35\textwidth}{!}{
\begin{tabular}{c|ccc}
\toprule\noalign{\smallskip}
\multirow {2}{*}{Method}  &\multicolumn{3}{|c}{Layout Pixel Error} \\
\cline{2-4}
 & overall & occluded & visible\\
\midrule
NYU Parser~\cite{silberman2012indoor} & 34.6 & 50.0 & 5.2\\
Guo et al.~\cite{guo2015predicting}& 10.9& $14.0$ &\textbf{4.8} \\
Ours& \textbf{10.6} & \textbf{13.6} & \textbf{4.8} \\
\bottomrule
\end{tabular}
}
\vspace{-2em}
\end{center}
\end{table}

\begin{table}
\begin{center}
\caption{Results of pixel labeling error for layout surfaces with 5 categories compared with Zhang et al.~\cite{zhang2013iccv} (on intersection of test subsets in \cite{silberman2012indoor} and \cite{zhang2013iccv}, 47 images). Note that~\cite{zhang2013iccv} models layout as boxes, while our method can model more general layouts }
\label{table:3d}
\resizebox{0.35\textwidth}{!}{
\begin{tabular}{c|ccc}
\toprule\noalign{\smallskip}
\multirow {2}{*}{Method}  & \multicolumn{3}{|c}{Layout Pixel Error}\\
\cline{2-4}
 & overall & occluded  & visible\\
\midrule
Zhang et al~\cite{zhang2013iccv} & 10.0 & 13.0 & 5.9\\
Guo et al.~\cite{guo2015predicting} & $5.4$ & $7.6$ & $2.3$\\

Ours& \textbf{5.1}& \textbf{7.2} & \textbf{2.0}\\

\bottomrule
\end{tabular}
}
\vspace{-2em}
\end{center}
\end{table}

\textbf{Layout estimation}. In Table~\ref{table:layout}, we evaluate accuracy of labeling background surfaces into ``left wall'', ``right wall'', ``front wall'', ``ceiling'', and ``floor''.  Ground truth is obtained by rendering the 3D annotation of layout surfaces, and our prediction is obtained by rendering our predicted layout surfaces.  The labels of ``openings'' (e.g., windows that are cut out of walls) are assigned based on the observed depth and surface normal. We compare to the RGBD region classifier of Silberman et al.~\cite{silberman2012indoor} on the full test set.  As expected, we outperform significantly on occluded surfaces (13.6\% vs. 50.0\% error) but also outperform on visible surfaces (4.8\% vs. 5.2\% error), which is due to the benefit of a structured scene model. Our method outperforms Guo et al.'s method on occluded surfaces, which means the layout estimation take benefits from the better object interpretation during the scene composition step. We also compare to Zhang et al.~\cite{zhang2013iccv} who estimate box-like layout from RGBD images on the intersection of their test set with the standard test set (Table.~\ref{table:3d}.  These images are easier than average, and our method outperforms substantially, cutting the error nearly in half (5.1\% vs. 10.0\%).

\begin{table}
\begin{center}
\caption{Results of freespace voxel estimation. Unoccupied voxel precision/recall using our method, given ground truth segmentation (GT-region) or only automatic region proposals. A baseline of freespace inferred from depth point (Sensor) is compared.}
\label{table:freespace}
\resizebox{0.30\textwidth}{!}{
\begin{tabular}{c|cc}
\toprule\noalign{\smallskip}
\multirow {2}{*}{Method}  & \multicolumn{2}{|c}{Freespace}\\
\cline{2-3}
 & precision & recall\\
\midrule
Sensor & $1.000$ & $0.7874$ \\
\midrule
Guo et al.~\cite{guo2015predicting} & $0.954$ & $0.914$ \\
Ours & $0.954$ &\textbf{0.919}\\
\midrule
%Guo et al.~\cite{guo2015predicting}, GT-region & $0.950$ & \textbf{0.931}\\
Ours w/ GT-region & 0.955& $0.925$ \\
\bottomrule
\end{tabular}
}
%\vspace{-2em}
\end{center}
\end{table}

\begin{table}
\begin{center}
\caption{Results of predicted object occupied voxel precision/recall,  compared to fitting bounding boxes to ground truth regions (Bbox).}
\label{table:occupancy}
\resizebox{0.40\textwidth}{!}{
\begin{tabular}{c|cccc}
\toprule\noalign{\smallskip}
\multirow {2}{*}{Method}  & \multicolumn{4}{|c}{Occupancy}\\
\cline{2-5}
& precision & recall & precision-$\epsilon$ & recall-$\epsilon$\\
\midrule
Bbox& 0.487& 0.298& 0.756 & 0.569\\
\midrule
Guo et al.~\cite{guo2015predicting}& \textbf{0.504} & $0.380$ & \textbf{0.751} & $0.646$ \\

Ours& $0.478$ & \textbf{0.397} & $0.741$ & \textbf{0.710}\\
\midrule

%Guo et al.~\cite{guo2015predicting}, GT-region &\textbf{0.575} & $0.364$ & \textbf{0.831} & 0.637\\

Ours w/ GT-region & 0.549 & 0.417& 0.815 & 0.681\\

\bottomrule
\end{tabular}
}
%\vspace{-2em}
\end{center}
\end{table}

\begin{table*}
\vspace{-1em}
\begin{center}
\caption{Comparison with Deep Sliding Shapes~(DSS)~\cite{Song_2016_CVPR}. We evaluate voxel occupancy~(our metric), semantic segmentation, and instance segmentation metric based on the predicted 19 object classes defined in DSS.}
\label{table:DSS}
\resizebox{0.75\textwidth}{!}{
\begin{tabular}{c|c|c|c|c|c|c|c|c|c}
\toprule\noalign{\smallskip}
\multirow {2}{*}{Method} & \multicolumn{4}{|c|}{Voxel occupancy} &  \multicolumn{3}{|c|}{Semantic segmentation} &  \multicolumn{2}{|c}{Instance segmentation} \\
\cline{2-10}
& precision & recall & precision-$\epsilon$ & recall-$\epsilon$ & avg class & avg instance & avg pixel & \begin{tabular}{c}Mean coverage \\ weighted\end{tabular} & \begin{tabular}{c}Mean coverage\\ unweighted\end{tabular}\\
\midrule
DSS\cite{Song_2016_CVPR}& 0.318 &\textbf{0.585} &0.572 & \textbf{0.667}& 22.92& \textbf{29.52} & \textbf{41.45}& 27.36&18.21\\
Ours & \textbf{0.381} & 0.275 & \textbf{0.642} & 0.490 & \textbf{24.72} & 27.57 & 33.64 & \textbf{33.75} & \textbf{25.57}\\

\bottomrule
\end{tabular}
}
\vspace{-1em}
\end{center}
\end{table*}

\begin{table*}
\vspace{-1em}
\begin{center}
\caption{Comparison with Semantic Scene Completion~(SSCNet)~\cite{song2016ssc}. We follow the metric used by SSCNet and evaluate on the NYUd v2 test split with kinect depth map.}
\label{table:SSC}
\resizebox{0.75\textwidth}{!}{
\begin{tabular}{c|ccc|cccccccccccc}
\toprule\noalign{\smallskip}
\multirow {2}{*}{Method} & \multicolumn{3}{|c|}{Scene completion} &  \multicolumn{12}{|c}{Semantic scene completion}\\
\cline{2-16}
& precision & recall & IoU & ceiling & floor & wall & window & chair & bed & sofa & table & tvs & furniture & objects & avg \\
\midrule
SSCNet & 57.0 & \textbf{94.5} & \textbf{55.1} & 15.1 & \textbf{94.7} & \textbf{24.4} & 0 & \textbf{12.6} & 32.1 & \textbf{35} &\textbf{13} & \textbf{7.8} & \textbf{27.1} & 10.1 & \textbf{24.7}\\
Ours & \textbf{69.9} & 63.0 & 49.4 & \textbf{19.4} &  68.2 & 21.0 & \textbf{16.5} & 10.7 & \textbf{43.1} & 22.2 & 0.7 & 4.3 & 23.0 & \textbf{15.3} & 22.8\\
\bottomrule
\end{tabular}
}
\vspace{-1em}
\end{center}
\end{table*}

\begin{table}
\vspace{-1em}
\begin{center}
\caption{3D Amodal object detection in NYUd v2 dataset. We evaluate on the mAP~(\%) as defined on~\cite{song2014sliding}}\label{tab:DSS_detec}
\resizebox{0.8\linewidth}{!}{
\begin{tabular}{c|ccccc}
\toprule\noalign{\smallskip}
Method & bed & toilet & sofa & table & chair \\
\midrule
Sliding Shapes~\cite{song2014sliding} & 33.5& 67.3 & 33.8 & 34.5 & 29\\
DSS~\cite{Song_2016_CVPR} &\textbf{84.7} & \textbf{89.9} & \textbf{55.4} & \textbf{70.5} & \textbf{61.1} \\
Ours & 36.9 & 36.1& 14.12 & 4.81 & 3.1 \\
\bottomrule
\end{tabular}}
\vspace{-1.5em}
\end{center}
\end{table} 

\begin{table*}
\vspace{-1em}
\begin{center}
\caption{Quantitative evaluation for our retrieval method compared with Guo et al's method.}
\label{table:retr}
\resizebox{0.6\textwidth}{!}{
\begin{tabular}{c|c|c|c|c|c|c|c|c|c}
\toprule\noalign{\smallskip}
\multirow {2}{*}{Method} & \multicolumn{3}{|c|}{avg class accuracy ($\%$)} &  \multicolumn{3}{|c|}{avg 3D IoU} &  \multicolumn{3}{|c}{avg surface distance (m)} \\
\cline{2-10}
& top 1 & top 2 & top 3 & top 1 & top 2 & top 3 & top 1 & top 2 & top 3\\
\midrule
Guo et al.~\cite{guo2015predicting} & $11.97$ & $16.36$ & $19.65$ & $0.134$ & $0.177$ & $0.200$ & $0.033$ &  $0.029$ & $0.027$\\
Ours& \textbf{41.56} &\textbf{54.47} &\textbf{62.07} & \textbf{0.191}& \textbf{0.231}&\textbf{0.249} & \textbf{0.026}& \textbf{0.024}&\textbf{0.023}\\

\bottomrule
\end{tabular}
}
\vspace{-1em}
\end{center}
\end{table*}

\begin{table*}
\begin{center}
\caption{\textbf{Our 81-class classification accuracy on ground truth 2D regions in the test set.} We compare two methods: our classification network with/without estimating support height. We compute the average accuracy for each class, average precision based on the predicted probability and the accuracy averaged over instances. The classification networks are trained and evaluated 10 times and the means and standard deviations (reflecting variation due to randomness in learning) are reported. 15 common object class results are also listed. Bold numbers signify better performance.}% compared with another method.}
\vspace{-1em}
\label{table:height}
\resizebox{1.0\textwidth}{!}{
\begin{tabular}{c|ccc|ccccccccccccccc}
\toprule\noalign{\smallskip}
Method & \begin{tabular}{c} avg per \\ class \end{tabular} & \begin{tabular}{c} avg  \\precision
\end{tabular} & \begin{tabular}{c} avg over\\instance
\end{tabular}  & picture & chair & cabinet & pillow & bottle & books & paper & table  & box & window & door & sofa & bag & lamp & clothes\\
\midrule
w/o support height&43.7$\pm$ 0.3 &37.7 $\pm$ 0.1& 40.8$\pm$ 0.3& 57.5& 46.1& 39.5& 66.5& \textbf{55.6}& 30.0 & 40.7&36.7&10.6 & 61.4& 54.9&63.0 & 14.9&64.6&25.3\\
w/ support height& \textbf{44.7}$\pm$ \textbf{0.3}&\textbf{39.7}$\pm$\textbf{0.1}& \textbf{42.7}$\pm$\textbf{0.2}& 57.5& \textbf{53.1} &\textbf{44.35} & \textbf{69.0}& 54.9&\textbf{33.5} &\textbf{43.5}&\textbf{39.5}&\textbf{14.1} & \textbf{62.4}& \textbf{57.8}&\textbf{65.9} & \textbf{15.1}&\textbf{65.9}&\textbf{26.9}\\
\bottomrule
\end{tabular}
}
\vspace{-1.5em}
\end{center}
\end{table*}

\begin{table}
\vspace{-1em}
\begin{center}
\caption{Classification accuracy for ground truth regions under different occlusion ratios in test set}\label{tab_occ}
\resizebox{0.6\linewidth}{!}{
\begin{tabular}{c|cc}
\toprule\noalign{\smallskip}
Occlusion Ratio  & $<0.5$ & $>0.5$ \\
\midrule
w/ support height & \textbf{45.8}& \textbf{38.5}\\
w/o support height &$44.2$ & $35.7$ \\
\bottomrule
\end{tabular}}
\vspace{-1.5em}
\end{center}
\end{table} 

%\begin{table}
%\vspace{-1em}
%\begin{center}
%\caption{Comparison with Semantic Scene Completion~(SSCNet)~\cite{song2016ssc} on the rendered NYU test set as~\cite{firman2016structured}}\label{tab:SSC_detec}
%\resizebox{0.8\linewidth}{!}{
%\begin{tabular}{c|ccc}
%\toprule\noalign{\smallskip}
%Method & precision & recall & IoU\\
%\midrule
%SSCNet completion & 66.3 & \textbf{96.9} & 64.8\\
%SSCNet joint & 75.0 &  92.3 &  \textbf{70.3} \\
%Ours & \textbf{76.1} & 59.9 & 50.3 \\
%\bottomrule
%\end{tabular}}
%\vspace{-1.5em}
%\end{center}
%\end{table} 

We evaluate layout depth prediction, the rendered depth of the room without foreground objects (Table~\ref{table:depth}). Error is the difference in depth from the ground truth layout annotation.  On visible portions of layout surfaces, the error of our prediction is very close to that of the sensor, with the difference within the sensor noise range.  On occluded surfaces, the sensor is inaccurate (because it measures the foreground depth, rather than that of the background surface), and the average depth error of our method is only 0.15 meters, which is quite good considering that the sensor noise is conservatively 0.03*depth.

%We perform two kinds of evaluations. First, we consider semantic labeling, labeling each pixel into {\em left wall, right wall, front wall, ceiling and floor}, a 5-way classification problem. The ground truth comes from rendering the annotation of the full 3D scene. We compare to directly using appearance-based features~\cite{silberman2012eccv}. For pixels that are labeled as ``openings'' in our prediction, we use the observed depth point to determine its type.
%
%Overall, our method has a clear advantage over pure appearance based methods in terms of labeling accuracy, shown in Figure~\ref{fig:layoutlabel}, with a $23.8\%$ absolute improvement in overall accuracy. It is relatively easy to predict surface labels when they are visible, with an error of $5.2\%$ just using appearance features. Our method further reduces it to $4.8\%$. The problem is much more challenging when layout surfaces are occluded; still, our method has an accuracy of $86.1\%$, much higher than that of appearance features. We also compare to Zhang et al.~\cite{zhang2013iccv}, and obtain $4.6\%$ better over all accuracy. Note that Zhang et al.~\cite{zhang2013iccv} used a subset of NYUv2~\cite{silberman2012eccv}, and we tested on the intersection of the test subset of~\cite{zhang2013iccv} and~\cite{silberman2012eccv}. Also, layout model in~\cite{zhang2013iccv} are boxes but our ground truth can have non-boxy layout.

%\subsection{Region Retrieval Evaluation}

\textbf{Occupancy and Freespace Evaluation.} We evaluate our scene prediction performance based on voxel prediction. The voxel representation has advantages of being computable from various volumetric representations, viewpoint-invariant, and usable for models constructed from multiple views (as opposed to depth- or pixel-based evaluations). The scope of the evaluation is the space surrounded by annotated layout surfaces. Voxels that are out of view or behind solid ground truth walls are not evaluated. We render objects separately and convert them into a solid voxel map. The occupied space is the union of all the voxels from all objects and layouts; free space is the complement of the set of occupied voxels.

Our voxel representation is constructed on a fine grid with $0.03m$ resolution to allow inspection of shape details of the 3D model objects we use.  The voxel prediction and recall are presented in Table~\ref{table:freespace} and Table~\ref{table:occupancy}.

There is inherent annotation and sensor noise in our data, which is often much greater than $0.03m$.  Objects, when they are small, of nontrivial shape, or simply far away, result in very poor voxel accuracy, even though they agree with the input image. Therefore, we perform evaluation with a tolerance, proportional to the depth of the voxel, for which we use $\epsilon=0.05*depth$, based on the sensor resolution of Kinect. Specifically, an occupied voxel within $\epsilon$ of a ground truth voxel is considered to be correct (for precision) and to have recalled that ground truth voxel.

We compare to two simple baselines. For free space, we evaluate the \textit{observed} free space from depth sensor. The free space from observed depth predicts 100\% of the visible free space but recalls none of the free space that is occluded.  For occupied space, our baseline generates bounding boxes based on ground truth segmentations with 10\% outlier rejection. We outperform this baseline, whether using ground truth segmentations or automatic region proposals to generate the model, and we perform similarly to the Guo et al. variation. Also, note that precision is higher than recall, which means it is more common to miss objects than to generate false ones. 

%\subsection{Instance Segmentation}

%We also evaluate the instance segmentation of our prediction based on 3D rendering of our predicted model (Fig.~\ref{fig:segmentationaccuracy}), following the protocol in RMRC~\cite{urtasun2013reconstruction}.  Even the ground truth annotations from Guo and Hoiem~\cite{guo2013iccv} (``3D GT'') do not achieve very high performance, because rendered models sometimes do not follow image boundaries well and some small objects are not modeled in annotations.  This provides an upper-bound on our performance.  We compare to the result of Gupta et al.~\cite{gupta2013cvpr}, by applying connected component on their semantic segmentation map.  %The result of their more recent paper~\cite{gupta2014eccv} is not available at this time. 
%Since our segmentation is a direct rendering of 3D models, it is more structured and expressive, but tends to be less accurate on boundaries, leading to loss in segmentation accuracy. Segmentation accuracy was not a direct objective of our work, and improving in this area is a possible future direction.

%\textbf{3D estimation.} We evaluate 3D estimation with layout pixel-wise and depth prediction and freespace and occupancy estimation. As shown in Fig.~\ref{table:3d},

\subsection{Comparison with Deep Sliding Shape}\label{sec:DSS}
We show in Table~\ref{table:DSS} the comparison between our approach and the DSS method. Our approach demonstrates better precision of voxel occupancy, average class semantic segmentation and both instance segmentation metrics, indicating a better shape representation based on CAD model over bounding box. DSS shows better recall in object voxel occupancy, which is benefited from their 3D local search regime for each object in the whole scene. The better recall of predicted 3D objects also lead to their better performance in the average pixel accuracy and average instance accuracy of semantic segmentation.

We compare our performance on 3D object detection metrics in Table~\ref{tab:DSS_detec}. Although 3D object detection is not our direct goal, our 3D detection on bed class is better than Sliding Shapes. We observe that our lower mAP for chair, sofa and table is due to three factors: 1) the failing of distinguishing between similar object classes: e.g. we predict on desk shape which is in ground truth a table, and we confuse between sofa and chair~(with arms like sofa shape); 2) 3D localization error, which is especially for toilet case; 3) our method is constrained to produce objects that do not overlap, even with classes outside the five main furniture. 

\subsection{Comparison with Semantic Scene Completion Network}\label{sec:SSC}

We show in Table~\ref{table:SSC} the comparison between our approach and the SSCNet method. For scene completion, our approach outperforms SSCNet in precision, which benefits from our detailed shape modeling for both objects and layout. SSCNet shows better performance in recall, due to their voxel-level prediction from observed occupancy space. Our lower recall also leads to our slightly lower performance in voxel-level occupancy IoU than SSCNet. For semantic scene completion, our predictions on ceiling, window, bed and other object classes outperform SSCNet. Our average voxel-level IoU is less than 2\% lower than SSCNet. Errors mainly come from objects like table with fewer observed depth points to fit object and shape classification confusion like tv~(classified as other objects) and sofa~(classified as bed or chair) class. 

%\subsection{Evaluation of region classification and shape retrieval}\label{subsec:evalretrieval}
\subsection{Ablation study}\label{subsec:evalretrieval}

\textbf{Region proposals.} In Tables~\ref{tab:ins} and ~\ref{table:sem}, we see the impact of changing the region proposal method from Guo et al.'s Prim's based algorithm to Gupta et al.'s~\cite{gupta2014learning} Multiscale Combinatorial Grouping (MCG).  The ``MCG + CCA'' method is Guo et al. with only a substitution of the region proposal method and improves over ``Guo et al.'' in each case.  We can also see that using ground truth regions improves performance over automatic region proposals, but the improvement is less than we had expected, which indicates that our region proposal and scene composition steps are effective.

%To demonstrate our advantages over Guo et al. gained from using improved region proposals, we evaluate the result of substituting Guo et al.'s region proposals with the region proposals used by our approach. Our approach utilize the the state-of-the-art RGBD region proposals by Gupta et al.~\cite{gupta2014eccv} based on Multiscale Combinatorial Grouping (MCG). We then perform non-maximal suppression and non-object class prunning to reduce the number of region proposals from 2000 in each image to 190 in each image, as explained in Sec.~\ref{CNN_retrieval}. We extract the region features same as Guo et al.'s approach from each of the region proposals. We then perform the same approach as Guo et al.'s that apply CCA retrieval to get top 10 shapes and then select the best 3 based on depth fitting. We name this method as ``MCG + CCA'' in Table~\ref{tab:ins} and Table~\ref{table:sem}. As we can see, by applying the MCG region proposals, the performance gets better in all the semantic segmentation evaluations, but still not reaching ours approach.

\begin{figure*}
\begin{center}
%\vspace{-1em}
%\fbox{\rule{0pt}{2.5in} \rule{0.9\linewidth}{0pt}}
   \includegraphics[width=0.9\linewidth]{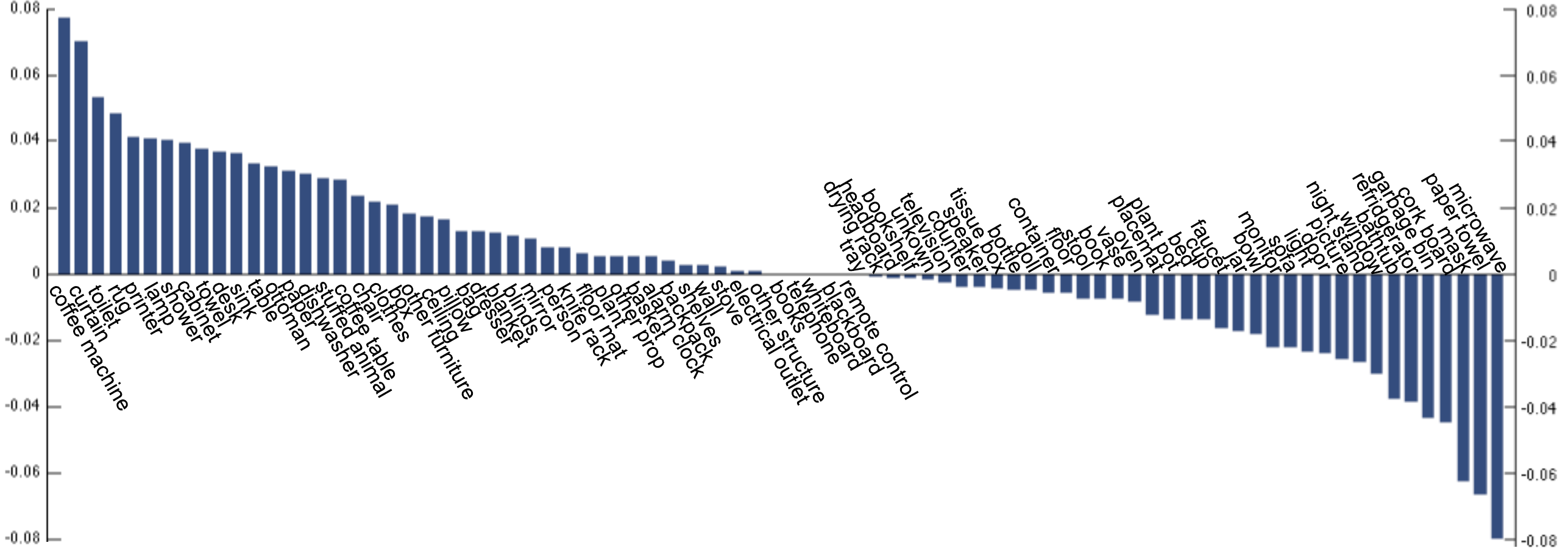}
\end{center}
\vspace{-1.5em}
   \caption{Semantic segmentation accuracy of each class w/ support minus accuracy w/o support accuracy with automatic region proposals}
\label{fig:semseg}
\end{figure*}

%\textbf{Categorization.} We report the region classification accuracy of our classification network on the groundtruth $2$D regions in the test set, as shown in Table~\ref{table:height}. Overall, incorporating support height prediction improves the classification results. For certain classes, objects often appear on the same height (e.g. chair, desk, rug) have better classification accuracy given object's height, while objects that can appear on several heights (e.g. picture) do not have this benefit. %This is conformed with our assumption that introducing object's support will help in classification of occluded regions.

%\textbf{PrediOcclusion reasoning.} 

%We also verify the effectiveness of occlusion reasoning by estimating object's support height. 

\textbf{Retrieval.} We evaluate our candidate shape retrieval method compared with Guo et al's retrieval method, as shown in table~\ref{table:retr}. To isolate from the influence of errors in other steps like region proposals generation or scene composition, we assume that the ground truth regions are given (and the candidate selection in Sec.~\ref{CNN_retrieval} for the CNN-based method is not performed). We evaluate top $N$ retrieved class accuracy and top $N$ retrieved shape similarity, based on the shape intersection over union (IoU) and the surface-to-surface distance~\cite{rock2015completing}. In our experiment, we set $N\in\{1,2,3\}$. To isolate from rotation ambiguity in shape similarity measurement, we rotate each retrieved object to find the best shape similarity score with the ground truth $3$D shape. Our CNN-based retrieval method outperforms the Guo et al.'s CCA method under all the evaluation criteria.

We can indirectly measure the effectiveness of our \textbf{scene composition} by comparing ``Ours'' to ``GT-Region'' (use the best fitting 3D object model and class for each ground truth 2D region).  For semantic segmentation, occupancy (Table~\ref{table:occupancy}), and free space (Table~\ref{table:freespace}, we see a relatively small improvement by using ground truth regions.  Qualititatively as well, our automatic region proposal and scene composition produce similarly good results to those produced using ground truth regions.  Although the region proposal method is far from perfect, our scene composition step appears to be effective at keeping the good region proposals and discarding the others.  

\textbf{Support height prediction.} We first examine the effect of predicting support height in region categorization. In table~\ref{table:height}, we report the region classification accuracy of our classification network given or without the support prediction on the ground truth $2$D regions in the test set. Overall, incorporating support height prediction improves the classification results. For certain classes, objects often appear on the same height (e.g. chair, desk, rug) have better classification accuracy given object's height, while objects that can appear on several heights (e.g. picture) do not have this benefit. %This is conformed with our assumption that introducing object's support will help in classification of occluded regions.
Table~\ref{tab_occ} shows the per-instance classification accuracy under different occlusion ratios of the ground truth regions in the test set. The occlusion ratio is computed from the observed 2D region over the 2D projection of the ground truth 3D label. The improvement in the classification accuracy is larger for highly occluded area, which agrees with our hypothesis that estimating object's support height will help classify occluded regions.

As an ablation study of the effect of incorporating support inference in the overall scene composition performance, in Table~\ref{tab:ins} and Table~\ref{table:sem}, we report our result with or without support height prediction and using the ground truth support height as the upper bound for including support reasoning. As we can see, incorporating support improves the semantic segmentation performance by about 1\%. Figure~\ref{fig:semseg} shows the effect of applying support reasoning on semantic segmentation accuracy of each class. Classes that are often at a certain height: toilet, printer, lamp, etc. will have benefits from support height information. Classes that might appear in several levels of height: mask, pixture, light won't gain benefits from this. Using ground truth support height generally leads to a larger benefit than estimated support height, as expected.

\section{Discussions}
We discuss the performance of our framework and the contributions of each of the component. Our framework follows three steps: region proposal, classification and shape retrieval, and scene composition. Each step solves for a loosely separate sub-problem, but are all closely related to produce a compact result for the complete 3D scene parsing task. For in-depth investigation of each of the component, as in the ablation study in Sec.~\ref{subsec:evalretrieval}, we perform experiments and analyze the effectiveness of the design choices in each step. For the region proposal step, we see improvements on applying the state-of-the-art RGBD region proposal method by Gupta et al.~\cite{gupta2014learning} instead of our previous Prim's based method. We observe less difference in overall performance if we use ground truth regions in our framework, which indicates the downstream step's robustness to errors from region proposals. For the retrieval step, our use of CNNs to classify and retrieve similar shapes outperforms our previous simpler CCA-based retrieval approach. We validate our hypothesis that estimating support height of objects can lead to better classification to improve retrieval, especially for occluded objects. The heavy occlusion characteristic of indoor scene makes the region classification harder, and accumulates errors to the retrieval step. To resolves the retrieval errors, we incorporate the scene composition step with combinatorial optimization to select a subset of shape candidates that fits the observation and scene regularity like no 3D overlap between shapes. We combine greedy search and a hill-climbing method to efficiently solve the hard optimization problem which outperforms a variety of other attempted methods including general integer programming algorithms and relaxations. Note that possible improvements for the framework exists, as we will discuss in the following paragraphs.%while as a novel approach to tackle the complex complete 3D scene parsing problem, in this paper we mainly emphasis on the integrity and the feasibility of our approach, and conduct extensive experiments to validate the framework performance.

Common errors made by our approach include splitting large objects into small ones, completely missing small objects, difficulty with highly occluded objects such as chairs. Based on the previous discussions, we see the major sources of error of our framework are: 1) region classification error and 2) shape fitting error. Errors mainly result from the heavy occlusion from objects in indoor scene. Fewer errors accumulate in the shape retrieval step, referring to the comparison between Table~\ref{table:height} and~\ref{table:retr}. %We also observe that applying various region proposal methods does not show obvious difference on the final performance, based on the comparison in Table~\ref{tab:ins},~ \ref{table:sem} and~\ref{table:occupancy}.
The method does not seem to be very sensitive to the region proposal method, based on the comparison in Table~\ref{tab:ins},~ \ref{table:sem} and~\ref{table:occupancy}. In fact, we were surprised to find that results from automatic region proposals are often comparable to those from ground truth regions. %, even though the automatic regions are often inaccurate. 
In part, this is because the later fitting and optimization steps are effective at discarding bad regions and retaining good ones. 

We see many interesting directions for future work: 1) modeling object context such as chairs tend to be near/under tables, relative pose among objects and the layouts; 2) modeling self-similarity, e.g. most chairs within one room will look similar to each other; 3) incorporating semantic constraints. For example, SSCNet by Song et al~\cite{song2016ssc} employs neural network to perform semantic completion from single depth image. %Our framework, instead of parametric modeling of scenes, performs a data-driven approach utilizing both RGB and depth information from single image. 
For further improvement, our system can incorporate the semantic completion from Song et al. to generate candidate region proposals and add semantic constraint for scene composition in Sec.~\ref{subsec:composition}.

\section{Conclusions} 
\label{sec:conclusion}
We proposed an approach to predict a complete 3D scene model of individual objects and surfaces from a single RGBD image. Our representation encodes the layout of walls as 3D planes and all interpretable objects as 3D mesh models, parsing both the visible and occluded portion of the scene. We take a data-driven approach, generating sets of potential object regions, matching to regions in training images, and transferring and aligning associated 3D models while encouraging fit to observations and spatial consistency. We incorporate support inference to aid interpretation and propose a retrieval scheme that uses CNNs to classify regions and find objects with similar shapes. We demonstrate the performance of our method on our newly annotated NYUd v2 dataset with detailed 3D shapes. Compared to our earlier work (Guo et al.~\cite{guo2015predicting}), we demonstrate improvements due to better region proposals and use of CNNs to classify and retrieve similar shapes.  We also show that our results from automatic region proposals are nearly as good as from ground truth regions, demonstrating the effectiveness of our scene composition and, more generally, of finding a scene hypothesis that is consistent in semantics and geometry.  We also showed that estimating support height of objects can lead to better classification, especially for occluded objects. 

\begin{acknowledgements}
%If you'd like to thank anyone, place your comments here
%and remove the percent signs.
This research is supported in part by ONR MURI grant N000141010934 and ONR MURI grant N000141612007. We thank David Forsyth for insightful comments and discussion and Saurabh Singh, Kevin Shih and Tanmay Gupta for their comments on an earlier version of the manuscript.
\end{acknowledgements}

% BibTeX users please use one of
%\bibliographystyle{spbasic}      % basic style, author-year citations
%\bibliographystyle{spmpsci}      % mathematics and physical sciences
%\bibliographystyle{spphys}       % APS-like style for physics
%\bibliography{}   % name your BibTeX data base

\bibliographystyle{spbasic}
\bibliography{thesisrefs}

% Non-BibTeX users please use
%\begin{thebibliography}{}
%
% and use \bibitem to create references. Consult the Instructions
% for authors for reference list style.
%
%\bibitem{RefJ}
% Format for Journal Reference
%Author, Article title, Journal, Volume, page numbers (year)
% Format for books
%\bibitem{RefB}
%Author, Book title, page numbers. Publisher, place (year)
% etc
%\end{thebibliography}

\end{document}